# MUG: Multi-human Graph Network for 3D Mesh Reconstruction from 2D Pose

Chenyan Wu[1] · Yandong Li[2] · Xianfeng Tang[3] · Zhe Huang[4] · James Z. Wang[1]

Received: 12 November 2024 / Accepted: 29 June 2026 / Published online: 5 August 2026

**Abstract**

Reconstructing multi-human body meshes from a single monocular image is a crucial yet challenging problem in computer vision. This problem requires not only generating individual body mesh models for each person but also estimating the relative 3D positions among subjects to produce a coherent scene representation. In this work, we introduce MUG (Multi-hUman Graph network), which employs a single graph neural network to construct coherent multi-human meshes using only 2D pose data as input. Our approach demonstrates that purely pose-based methods can effectively perform simultaneous depth reasoning and multi-human mesh generation. Existing image-based methods typically rely on lab-collected training datasets with accurate 3D labels; however, these datasets often introduce an image domain gap when applied to in-the-wild testing data or art images due to differences in appearance and context. In contrast, MUG leverages the geometric consistency of 2D poses across diverse datasets, mitigating domain discrepancies. The MUG network operates in three primary phases. Initially, to model the multi-human environment, it processes multi-human 2D poses and constructs a novel heterogeneous graph. This graph connects nodes both across different people and within individuals to capture inter-human interactions and accurately represent body geometry, including skeletal and mesh structures. Subsequently, it employs a dual-branch graph neural network: one branch predicts inter-human depth relations, while the other predicts the root-joint-relative mesh coordinates. Finally, the complete multi-human 3D meshes are constructed by combining the outputs from both branches. Despite the simplicity of using only 2D pose inputs and streamlined network architecture, MUG outperforms existing multi-human mesh estimation methods. This superiority is consistently observed across various datasets from diverse domains, such as RH, MuPoTS-3D, and 3DPW. Qualitative results show that MUG can effectively handle art images, through-wall scenarios, and poor lighting conditions when incorporating advanced 2D pose networks. Both qualitative and quantitative evaluations highlight MUG's remarkable generalization ability in open-world scenarios.

## 1 Introduction

Deep learning methods leveraging large-scale datasets have shown tremendous progress in tasks involving 3D human mesh recovery, which is often a crucial initial step in computer vision and robotics applications focusing on human

✉ Chenyan Wu
czw390@psu.edu

Yandong Li
lyndon.leeseu@outlook.com

Xianfeng Tang
tangxianfeng@outlook.com

Zhe Huang
iamzhehuang@gmail.com

James Z. Wang
jwang@psu.edu

1 Department of Informatics and Intelligent Systems, College of Information Sciences and Technology, The Pennsylvania State University, University Park, PA 16802, USA

2 Google Research, Bellevue, WA 98004, USA

3 Amazon, Palo Alto, CA 94301, USA

4 The Robotics Institute, Carnegie Mellon University, Pittsburgh, PA 15213, USA

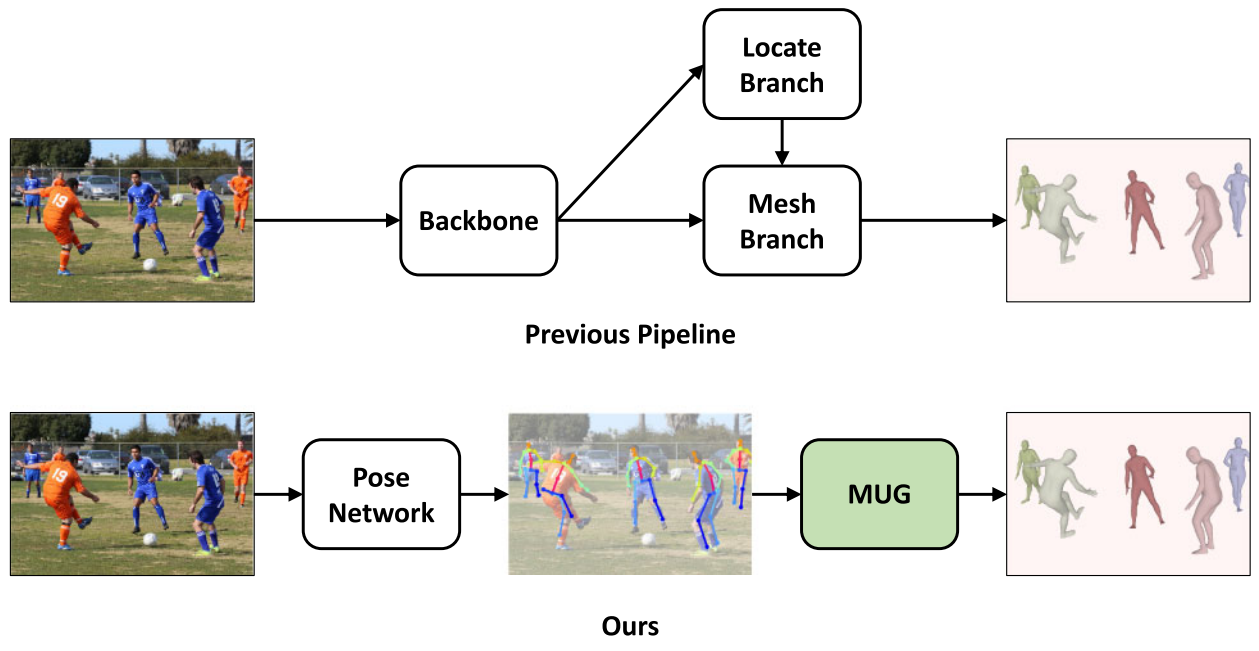

**Fig. 1** Comparison of existing multi-human mesh networks with our Multi-hUman Graph (MUG) network. Traditional networks typically adopt a detection-style pipeline using images as input. In contrast, the MUG takes 2D pose data as input, which can be obtained by 2D human pose estimation networks.

behavior, such as activity recognition (Ke et al., 2013) and bodily expressed emotion understanding (Luo et al., 2020; Wu et al., 2023; Wang et al., 2023). Notable achievements in single-human mesh estimation have been reported in recent studies (Kanazawa et al., 2018; Kolotouros et al., 2019b, a; Choi et al., 2020; Moon & Lee, 2020; Zheng et al., 2022a; Lin et al., 2021; You et al., 2023). Typically, these methods are categorized based on their input type: pose-based approaches that rely on 2D human poses (Choi et al., 2020; Zheng et al., 2022a; You et al., 2023), and image-based approaches that utilize images as input (Kanazawa et al., 2018; Kolotouros et al., 2019b, a; Moon & Lee, 2020; Lin et al., 2021).

Given the successes achieved in single-human scenarios, it is a natural progression to extend these methods to multi-human scenarios. However, in the domain of multi-human mesh estimation—which requires generating body mesh models and providing comprehensive 3D positions for each instance—image-based approaches have been predominant. Most recent methods (Jiang et al., 2020; Zhang et al., 2021; Sun et al., 2021, 2022) adopt a detection-style pipeline, taking entire images as input. This process is illustrated in Fig. 1, where images are initially input into a backbone network to extract features. One branch of this network locates human instances, while another branch (or multiple branches) generates the body meshes, utilizing both the image feature and instances' locations. Particularly, Jiang et al. (2020) leveraged the region proposal network (RPN) (Ren et al., 2015) for proposing human body boxes, while Zhang et al. (2021) and Sun et al. (2021, 2022) identified body center points in their locating branches.

However, these image-based approaches suffer from the domain gap between natural settings (*i.e.*, "in the wild") and controlled laboratory environments where most 3D datasets are collected. This issue is a well-documented concern in 3D-related research (Zhou et al., 2017; Wang et al., 2019a; Choi et al., 2020; Gong et al., 2021; Martinez et al., 2017; Choi et al., 2022). This largely reflects reliance on lab-collected datasets–such as Human3.6M (Ionescu et al., 2013) and HumanEva (Sigal et al., 2010)–for accurate 3D supervision. Although in-the-wild 2D data (e.g., COCO) can be incorporated, precise 3D labels are predominantly lab based, which can limit generalization to domains not represented in training. Lab imagery also differs markedly from test-time data in background, lighting, and occlusion statistics: laboratory images are relatively uniform, whereas in-the-wild scenes are diverse and cluttered.

This study introduces a pose-based pipeline for multi-human mesh estimation that helps mitigate the domain gap in this area. 2D poses provide a compact, geometry-centric representation that transfers more consistently across datasets than raw image features: regardless of lighting, weather, or style, they are simply the pixel coordinates of anatomical keypoints. High-quality 2D poses are readily available from state-of-the-art keypoint detectors (Sun et al., 2019; Cao et al., 2017; Xu et al., 2022). Our pipeline is detector-agnostic and can immediately benefit from stronger pose estimators; moreover, it can ingest keypoints from non-visual modalities (e.g., radio), reducing sensitivity to image-appearance domain shift.

However, reconstructing *multiple* human meshes from pose data alone raises two central challenges that are distinct from single-human settings.

- *Estimating relative depth from 2D poses*: Can 2D poses alone support reliable 3D placement and depth ordering of multiple people in the same scene? Unlike single-human methods that focus on reconstructing an individual's body mesh relative to their root joint, multi-human reconstruction must accurately determine the relative 3D positions among multiple subjects within the same scene.
- *Unified network for simultaneous depth and multi-human mesh output*: Can a single unified network jointly estimate inter-person 3D relationships while producing a mesh for each subject in one pass? This entails generating multiple human meshes simultaneously—a complex task that extends well beyond single-subject reconstruction.

While Pose2Mesh (Choi et al., 2020) demonstrates single-person mesh reconstruction from poses, it does not address inter-person localization. Likewise, 3DCrowdNet (Choi et al., 2022) incorporates 2D poses in crowded scenes but primarily relies on image features and processes individuals sequentially rather than jointly. These requirements are particularly demanding for a purely pose-based approach and have largely been overlooked in prior work.

In this work, we propose a simple yet effective heterogeneous graph convolutional network named MUG (Multi-hUman Graph network). The complete framework of our approach is illustrated in Fig. 1. This new pipeline incorpo-rates several key design elements that enable us to address the two aforementioned challenges:

- *Heterogeneous graph architecture:* This graph differentiates between two types of nodes: body joints and mesh vertices. Within each individual human, joint and mesh nodes are interconnected to form intra-human edges, aligning with the anatomical structure of the body skeleton and the mesh topology. To represent inter-human relations, we selectively establish edges between appropriate joint nodes across different humans. Such design effectively captures both intra- and inter-human relationships, enabling accurate prediction of relative depths and human meshes.
- *Feature construction:* We carefully design the feature input to optimize reconstruction accuracy. Particularly, we utilize absolute poses rather than normalized poses, which significantly enhances depth estimation performance. We also explore additional informative features to further improve mesh generation accuracy.
- *Dual-branch GCN:* Our dual-branch graph convolutional network architecture is designed to simultaneously output root-camera depth and root-relative body mesh coordinates for multiple humans. After feeding the heterogeneous graph and the 2D pose data into this neural network, we can construct multi-human meshes by combining the outputs from both branches.

Extensive experiments were conducted on a broad spectrum of datasets to validate the superior capabilities of MUG. Our results demonstrate that MUG outperforms the state-of-the-art (SOTA) multi-human methods on the RH dataset (Sun et al., 2022), affirming its proficiency in 3D depth reasoning. Under cross-dataset testing conditions, MUG exceeded the performance of the former SOTA on both MuPoTS-3D (Mehta et al., 2018) and 3DPW (Von Marcard et al., 2018), evidencing its exceptional generalization ability. Additionally, a wide array of qualitative examples is provided to further demonstrate our approach's adaptability to diverse image types. Even in challenging scenarios–such as art images, through–wall imaging, and poor lighting conditions-where SOTA image-based methods may fail due to domain gaps, our method continues to perform effectively. The paper also includes comprehensive ablation studies for a more in-depth analysis.

Our **contributions** are summarized as follows:

- *Methodological innovation:* We have developed a new pipeline for multi-human reconstruction. This approach demonstrates that using 2D pose as input, a singular graph network can simultaneously output both multi-human meshes and their 3D locations, a capability rarely seen in previous purely pose-based work.
- *New network architecture:* We introduce MUG, a novel graph neural network (GNN) that leverages the heterogeneous graph composed of joint nodes, vertex nodes, and different types of edges to effectively represent both intra- and inter-human relations.
- *Empirical generalization performance:* Our results indicate that using poses only reduces sensitivity to appearance-domain shift and yields strong multi-person performance in the wild, while remaining competitive on 3DPW and Panoptic. We do not claim universal superiority over image-feature methods; rather, our approach offers complementary strengths–trading small differences on some joint-based metrics for notable robustness under pronounced domain shifts.

In this paper, we present a purely pose-based approach for multi-human mesh estimation that utilizes minimal input information. While RGB images provide more comprehensive data, we demonstrate that relying solely on pose data can effectively handle multi-person scenarios and perform competitively with image-based methods across several datasets, particularly in terms of depth reasoning and mesh reconstruction. We hope that our design will serve as a solid baseline and inspire further research in multi-human mesh estimation.

## 2 Related Work

We present an overview of related literature. The references are illustrative examples and do not constitute an exhaustive survey of the entire domain.

### 2.1 Single-human Mesh Estimation

Single-human mesh estimation aims at reconstructing a body mesh from a cropped image of a human subject. The body mesh can be represented by 3D vertex coordinates or parametric human body models such as the Skinned Multi-Person Linear (SMPL) model (Loper et al., 2015). As mentioned earlier, human mesh estimation methods can be categorized into image-based and pose-based approaches based on their input.

Many developments in this field have adopted end-to-end neural networks to train images and regress SMPL parameters. Kanazawa et al. (2018), with their HMR system, utilized adversarial loss in their regression network. Pavlakos et al. (2018) exploited 2D joint heatmaps and silhouettes as cues. Kolotouros et al. (2019a), in their work on SPIN, introduced optimization techniques into the regression network. Recent studies have mainly used image-based pipelines to output 3D vertex coordinates. GraphCMR (Kolotouros et al., 2019b) used ResNet to extract global features, followed by a GNN to regress 3D vertex coordinates. Inspired by heatmap representations in 2D pose estimation, Moon and Lee (2020) developed a new representation named “lixel” to aid in regressing 3D coordinates. Lin et al. (2021) developed a model leveraging a Transformer network.

Bogo et al. (2016) pioneered a pose-based method, demonstrating a way to fit SMPL parameters by minimizing the error between the predicted and input pose. Pose2Mesh (Choi et al., 2020) takes the 2D pose as input, exploits the 3D pose as intermediate features, and uses a GNN to estimate body mesh coordinates. Our work also uses 2D pose as input and adopts GNN. However, unlike single-human problems, we need to estimate human depth and analyze inter-human interactions. The method by Choi et al. (2020) only outputs root-relative coordinates; therefore, when reprojecting body meshes onto the original image, it cannot distinguish the order of multiple human bodies. Subsequent pose-based approaches, including Zheng et al. (2022a) and You et al. (2023), mainly adopted Transformer networks rather than GNNs. 3DCrowdNet (Choi et al., 2022) leverages 2D pose heatmaps to construct robust image features, facilitating the recovery of a single person’s 3D human mesh from crowded environments.

### 2.2 Multi-human Mesh Estimation

Multi-human mesh estimation is a relatively new field (Zanfir et al., 2018a, b; Jiang et al., 2020; Zhang et al., 2021; Sun et al., 2021). A pioneering work by Zanfir et al. (2018a) implemented a top-down strategy. First, a multi-task single-human network was used to estimate 2D pose, 3D mesh, and other scene constraints. These outputs were then jointly optimized to generate the final prediction. In another work, Zanfir et al. (2018b) proposed a bottom-up method, starting with 3D pose estimation and then utilizing an unsupervised auto-encoder to estimate SMPL parameters.

More recent approaches have adopted detection-style pipelines to solve this task and achieve stunning performance. Jiang et al. (2020) followed the two-stage detection framework Faster RCNN (Ren et al., 2015) to use RPN to propose human body boxes, followed by an SMPL head to regress SMPL parameters for these proposals. Zhang et al. (2021) and the ROMP project by Sun et al. (2021) adopted a one-stage solution, using one head to locate the center point of the human subject and another to output SMPL parameters relative to this center point. However, as previously mentioned, these methods suffer from the issue of domain shift.

### 2.3 GNNs

GNNs learn node embeddings that preserve the graph structure and the original node attributes (Kipf & Welling, 2016). Recently, GNNs have achieved SOTA performance on various computer vision tasks (Li et al., 2019; Choi et al., 2020; Yang et al., 2020). Based on the type of graphs they operate on, GNNs can be divided into two categories: homogeneous GNNs (Kipf & Welling, 2016; Veličković et al., 2017; Hamilton et al., 2017) and heterogeneous GNNs (Schlichtkrull et al., 2018; Wang et al., 02019). Homogeneous GNNs operate under the assumption that all nodes and edges within the graph are of a uniform type. This category primarily focuses on modeling such relations via different approaches, including GCNs (Kipf & Welling, 2016), attention networks (Veličković et al., 2017), and recurrent neural networks (Hamilton et al., 2017). On the other hand, heterogeneous GNNs are designed for graphs comprising different types of nodes and edges. The diversity necessitates GNNs to capture the distinct characteristics of each node and edge type. A notable example is the R-GCN (Schlichtkrull et al., 2018), which uses individual GCNs (Kipf & Welling, 2016) for each type of node and edge.

## 3 Method

This section details our technical approach. The entire framework is illustrated in Fig. 2. We describe our proposed heterogeneous multi-human graph in Section 3.1. This is followed by an in-depth explanation of the node feature construction in Section 3.2. The architecture of the neural network is described in Section 3.3. In Section 3.4, we analyze the computation of root-camera depth. Finally, Section 3.5 outlines the process of reconstructing multiple human figures using the outputs generated by the network.

### 3.1 Graph Construction

We introduce the methodology for constructing the graph representation of the MUG, as illustrated in Fig. 3. Considering a scenario involving $K$ individuals, we use $\mathcal{G}^*$ to denote the entire multi-human graph and $\mathcal{G}_1, \cdots, \mathcal{G}_K$ to denote the subgraphs corresponding to each person. Each subgraph contains two types of nodes: joint nodes, denoted as $\mathcal{V}_i^{\text{Joint}}$, and mesh nodes, denoted as $\mathcal{V}_i^{\text{Mesh}}$.

The joint nodes correspond to human body keypoints, such as hips and shoulders, with a total of $J$ joint nodes per individual. These nodes are characterized by their 2D coordinate ground truth in training datasets and can be accurately predicted by 2D pose detection networks including He et al. (2017); Sun et al. (2019); Xu et al. (2022). The mesh nodes represent the vertices of the body mesh, with a total of $V$ mesh nodes per individual. Our objective is to obtain their absolute 3D coordinates.

To represent the geometric topology of the human body within each individual subgraph, we establish connections between joint nodes according to the body skeleton structure, and we link mesh nodes based on the body mesh topology.

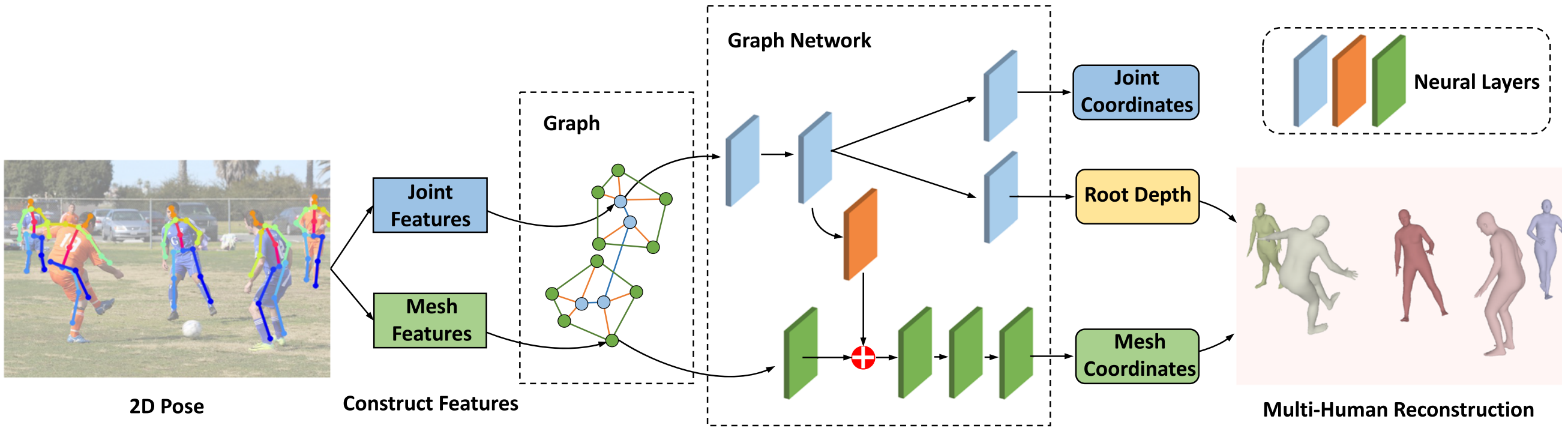

**Fig. 2** Illustration of the MUG framework. The input to the system is the 2D multi-human pose. To explicitly model both inter- and intra-human relations, a heterogeneous graph containing joint nodes (blue circles) and mesh nodes (green circles) is built. For a detailed illustration of the graph structure, please refer to Fig. 3. On top of the graph, we construct node features using the 2D pose. The graph network adopts a dual-branch structure: joint nodes with joint features are processed through the upper branch (illustrated with blue blocks), while mesh nodes carrying mesh features are directed through the lower branch (indicated by green blocks). A GCN layer (represented by an orange block) connects these two branches. Finally, the upper branch outputs root-camera depth, and the lower branch computes root-relative mesh coordinates. By combining the outputs from these branches, the full multi-human meshes are reconstructed. Additionally, the upper branch furnishes joint coordinates, which contribute to the learning process.

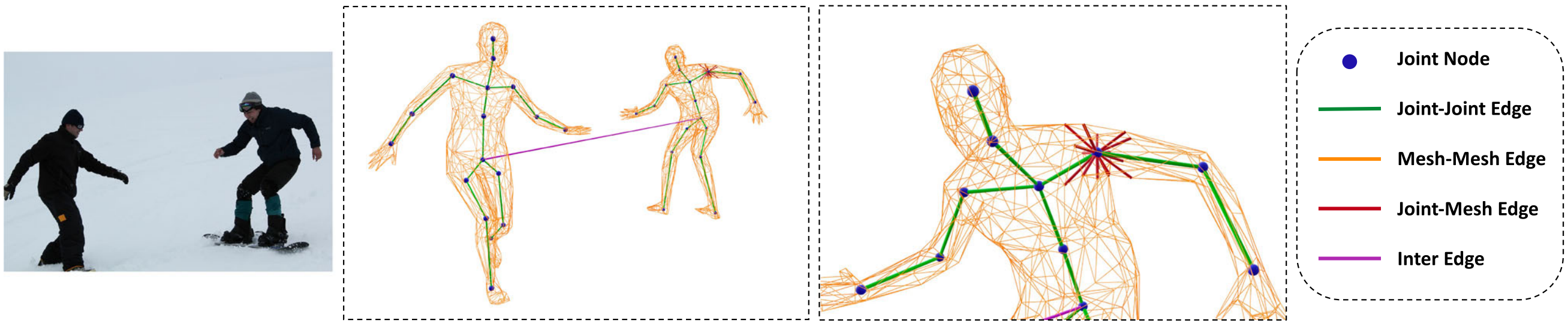

**Fig. 3** Structure of the graph. The graph consists of two kinds of nodes, joint nodes $\mathcal{V}^{\text{Joint}}$ (blue circles) and mesh nodes $\mathcal{V}^{\text{Mesh}}$. Each human corresponds to one subgraph of the whole graph. Within one human, joint nodes are connected to get edge $\mathcal{E}^{\text{Joint - Joint}}$ (green lines) according to body skeleton structure; following body mesh topology, mesh nodes are connected to get edge $\mathcal{E}^{\text{Mesh - Mesh}}$ (orange lines); each mesh node connects to its two nearest joint nodes, generating edge $\mathcal{E}^{\text{Joint - Mesh}}$ (red lines). To represent inter-human relations, we connect joint nodes across humans to get edge $\mathcal{E}^{\text{Inter}}$ (purple lines). Panels (b) and (c) show the graph structure for image (a). For clarity, mesh nodes and some edges of $\mathcal{E}^{\text{Joint - Mesh}}$ are not displayed.

Furthermore, we add edges between these two types of nodes according to their distance as defined by the SMPL body mesh template (Loper et al., 2015); specifically, each mesh node is linked to its two nearest joint nodes. These edges are unidirectional, extending exclusively from joint nodes to mesh nodes, reflecting the fact that the input for MUG comprises only the 2D coordinates of joints. Therefore, coordinate information is transmitted strictly from the joint nodes to the mesh nodes.

Formally, each subgraph $\mathcal{G}_i$, where $i \in \{1, \ldots, K\}$, is represented as $\mathcal{G}_i = (\mathcal{V}_i, \mathcal{E}_i, \mathbf{X}_i)$. The node set $\mathcal{V}_i = \mathcal{V}_i^{\text{Joint}} \cup \mathcal{V}_i^{\text{Mesh}}$, the edge set $\mathcal{E}_i = \mathcal{E}_i^{\text{Joint - Joint}} \cup \mathcal{E}_i^{\text{Mesh - Mesh}} \cup \mathcal{E}_i^{\text{Joint - Mesh}}$, and the node features $\mathbf{X}_i = \mathbf{X}_i^{\text{Joint}} \cup \mathbf{X}_i^{\text{Mesh}}$. The construction of these node features is detailed in Section 3.2.

To model the spatial correlation among multiple individuals, we integrate all human subgraphs into a unified multi-layer graph, denoted as $\mathcal{G}^* = (\mathcal{V}^*, \mathcal{E}^*, \mathbf{X}^*)$. Each layer within this graph corresponds to a human subgraph $\mathcal{G}_i$, so we have $\mathcal{V}^* = \{\mathcal{V}_i\}_{i=1}^K$, and $\mathbf{X}^* = \{\mathbf{X}_i\}_{i=1}^K$. We construct edges $\mathcal{E}^{\text{Inter}}$ across different subgraphs to represent the inter-human relations. Specifically, we connect the root-joint nodes of every pair of individuals. The root joint usually serves as a critical marker for determining the depth of a human relative to the camera's position; thus, relative depth information for each human can be propagated through these root-to-root edges. A detailed analysis of $\mathcal{E}^{\text{Inter}}$ is provided in the ablation study (Section 4.5).

In summary, the proposed graph model incorporates intra-human edges within each human subgraph—including $\mathcal{E}^{\text{Joint - Joint}}$, $\mathcal{E}^{\text{Joint - Mesh}}$, and $\mathcal{E}^{\text{Mesh - Mesh}}$—as well as $\mathcal{E}^{\text{Inter}}$ that connect different individuals. The complete set of edges is described as $\mathcal{E}^* = \mathcal{E}^{\text{Inter}} \cup \{\mathcal{E}_i\}_{i=1}^K$.

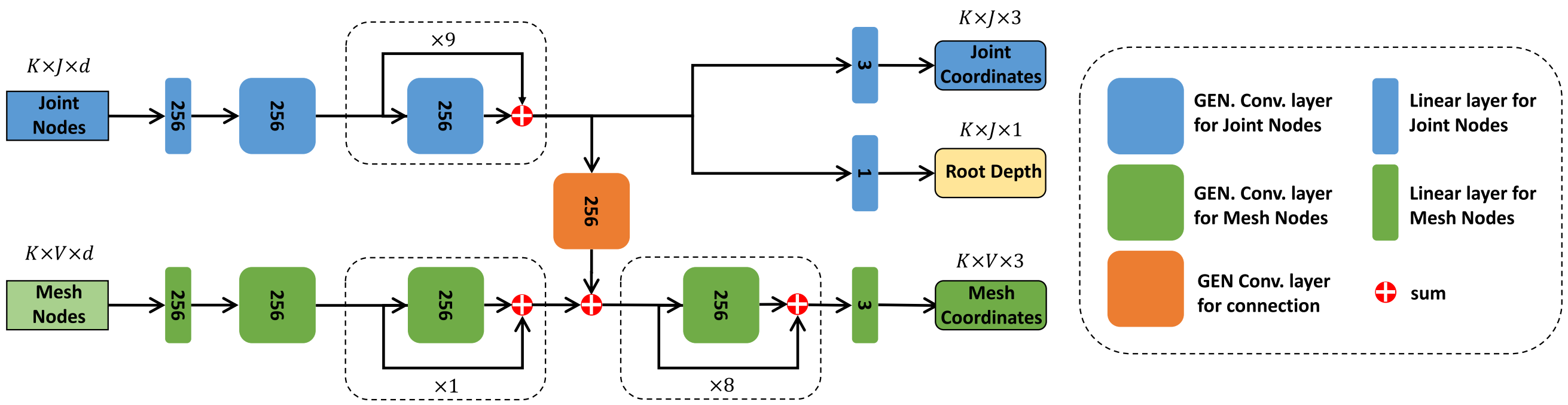

**Fig. 4** Detailed network structure. The number inside each neural network layer represents the feature dimension of the layer output.

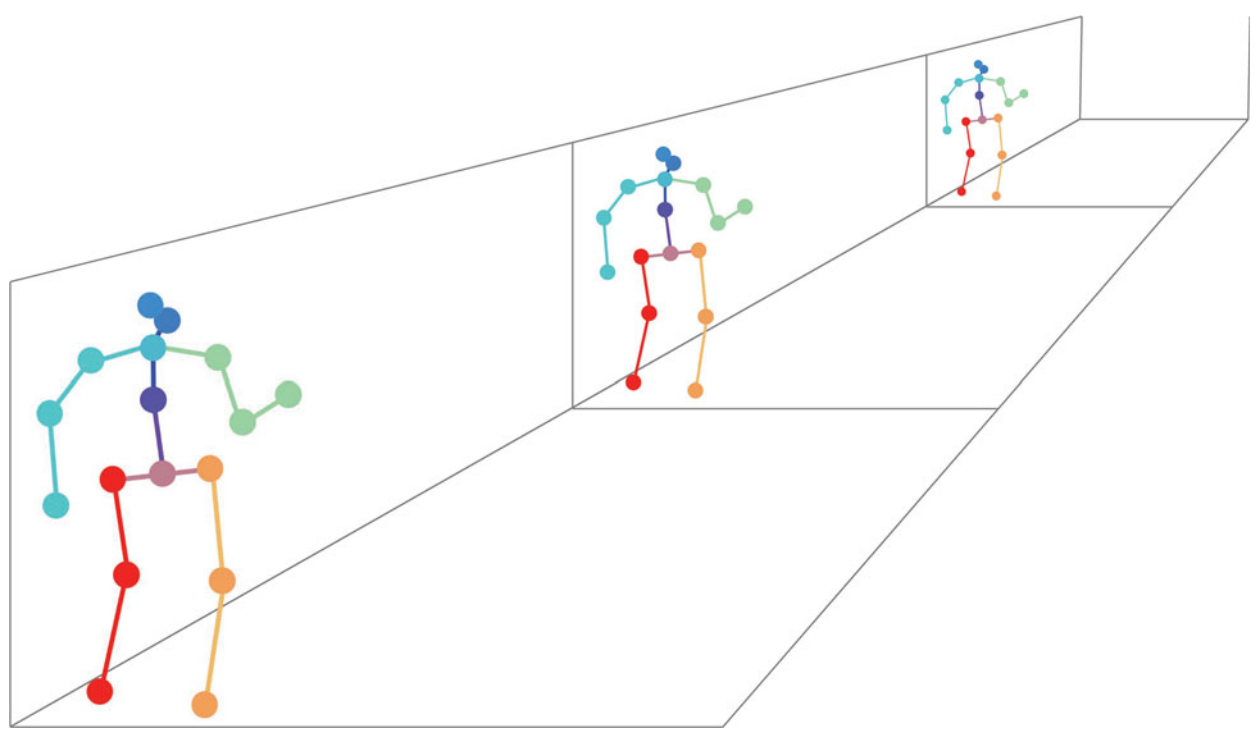

**Fig. 5** Normalized human poses alone cannot deduce 3D positioning. After normalization, the three human poses in the image share identical joint coordinates, preventing a neural network from discerning depth. Absolute joint coordinates are essential for accurate 3D position inference.

## 3.2 Node Feature Construction

We define the set of node features as $\mathbf{X}^* = \{\mathbf{X}_i\}_{i=1}^K$, where $\mathbf{X}_i$ represents the features of the $i^{\text{th}}$ subject. These features comprise two components: $\mathbf{X}_i^{\text{Joint}} \in \mathbb{R}^{J\times d}$ for joint nodes and $\mathbf{X}_i^{\text{Mesh}} \in \mathbb{R}^{V\times d}$ for mesh nodes. Here, $d$ denotes the dimensionality of the feature space.

For joint nodes, we denote the feature for the $j^{\text{th}}$ joint of the $i^{\text{th}}$ human as $\mathbf{X}_{ij}^{\text{Joint}}$. The primary information used to construct these features is the 2D pose input, denoted as $P_i^{\text{2D}} \in \mathbb{R}^{J\times 2}$. Following previous studies (Wandt & Rosenhahn, 2019; Choi et al., 2020), we first apply a standard normalization process to $P_i^{\text{2D}}$. Specifically, we subtract the mean of $P_i^{\text{2D}}$ and then divide it by the standard deviation to obtain the normalized pose $\overline{P}_i^{\text{2D}}$. Then normalized coordinate $\overline{P}_{ij}^{\text{2D}}$ is assigned to the $j^{\text{th}}$ joint.

While this normalization process (Wandt & Rosenhahn, 2019) allows the convergence of the GCN, it transforms each human's 2D pose into relative coordinates, disregarding individual body size and position. To analyze relative 3D human positions, we need to compare the absolute joint coordinates across humans in the images. Thus, the joint coordinates over the entire image range are necessary to pass to joint nodes. Fig. 5 provides a visual explanation.

To facilitate training and ensure consistency, we resize the raw image's longer dimension to 1000 pixels, centering the image within a 1000 × 1000 pixels square. The pose $P_i^{\text{2D}}$ is accordingly transformed into $\widetilde{P}_i^{\text{2D}}$. The transformed coordinate $\widetilde{P}_{ij}^{\text{2D}}$ is assigned to the $j^{\text{th}}$ joint. To fully utilize $\widetilde{P}_i^{\text{2D}}$, we also pass the entire $\widetilde{P}_i^{\text{2D}}$ to $\mathbf{X}_{ij}^{\text{Joint}}$.

In addition, we leverage 11 SMPL body mesh models, containing the SMPL template model as well as 10 randomly selected from the Human3.6M dataset. Mesh vertex coordinates can be computed from these models. For conciseness, we denote the mesh coordinates from one model as $T \in \mathbb{R}^{V\times 3}$. We compute the 3D coordinates of joints by multiplying $T$ with the joint regression matrix $\mathcal{J} \in \mathbb{R}^{J\times V}$, where $\mathcal{J}$ is defined in the SMPL body mesh model (Loper et al., 2015). Then we pass $(\mathcal{J}T)_j$ to $\mathbf{X}_{ij}^{\text{Joint}}$. Although these body mesh coordinates are fixed, the GNN can learn some rigid geometry information by utilizing these examples.

Finally, the joint feature is obtained by concatenating the aforementioned features,

$$\mathbf{X}_{ij}^{\text{Joint}} = (\widetilde{P}_{ij}^{\text{2D}}, \widetilde{P}_i^{\text{2D}}, \overline{P}_{ij}^{\text{2D}}, (\mathcal{J}T)_j)\,. \tag{1}$$

For mesh nodes, we denote the feature for the $v^{\text{th}}$ mesh vertex in $i^{\text{th}}$ human as $\mathbf{X}_{iv}^{\text{Mesh}}$. Unlike joint nodes, we cannot determine the $2D$ position information of each mesh vertex directly from the 2D pose input. We directly use $\widetilde{P}_{ij^*}^{\text{2D}}$ and $\overline{P}_{ij^*}^{\text{2D}}$ to initialize the mesh features, where $j^*$ represents the joint nearest to the $v^{\text{th}}$ mesh node in the SMPL body mesh template. Following a similar process with joint nodes, we also feed $\widetilde{P}_i^{\text{2D}}$ and $T_v$ (the $v^{\text{th}}$ vertex in $T$) to $\mathbf{X}_{iv}^{\text{Mesh}}$. By concatenating these features, we obtain,

$$\mathbf{X}_{iv}^{\text{Mesh}} = (\widetilde{P}_{ij^*}^{\text{2D}}, \widetilde{P}_i^{\text{2D}}, \overline{P}_{ij^*}^{2D}, T_v)\,. \tag{2}$$

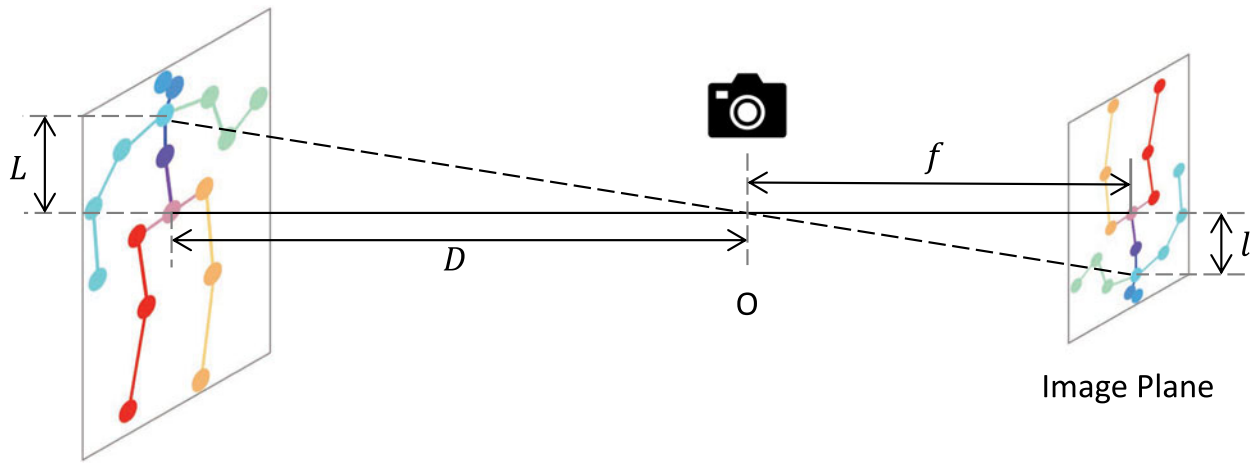

**Fig. 6** Visualization of a pinhole camera model. $L$ and $l$ denote the actual length of the spine and its length in the image, respectively.

## 3.3 Network Structure

The architecture of our model, illustrated in Fig. 4, incorporates a dual-branch graph network to process two types of nodes. The first branch processes joint nodes endowed with joint features, while the second branch handles mesh nodes equipped with mesh features. These branches are interconnected via a GCN layer, referred to as the connection layer.

In the joint nodes branch, processing initiates with a linear layer, followed by a GCN layer and a series of nine consecutive GCN blocks. The output from this branch then diverges into three distinct pathways. The first pathway integrates with the mesh branch via the connection layer. The second pathway passes through an additional linear layer, effectively reducing the feature dimensions to 1. This dimension-reduced feature is averaged over all joint nodes to provide a depth estimate for each human instance. The third pathway, also processed through a linear layer, reduces dimensions to 3 to produce 3D joint coordinates. Although these coordinates do not contribute directly to 3D mesh reconstruction, they are instrumental in guiding the branch towards learning depth representation.

Similarly, the mesh nodes branch starts with one linear layer, followed by a GCN layer and nine consecutive GCN blocks. After the initial GCN layer and one GCN block, the mesh nodes combine their features with the output from the connection layer. Subsequently, they undergo processing through eight more GCN blocks, eventually reducing the feature dimension to 3. This process culminates in the output of root-relative body mesh coordinates.

The GCN blocks used are GCN residual blocks, as proposed by DeeperGCN (Li et al., 2020), enhancing the robustness and depth of our model's processing capabilities. All GCN layers are implemented as GENeralized Graph Convolution layers.

## 3.4 Estimating Depth

Existing multi-human pose or mesh estimation methods, such as HMOR (Wang et al., 2020), SMAP (Zhen et al., 2020), and BEV (Sun et al., 2022), use various image features to learn monocular depth. In this section, we demonstrate how to use only 2D poses for monocular depth reasoning.

In the context of monocular imaging, predicting a precise root-camera depth is not feasible. Instead, our objective is to provide a comprehensive depth estimation that reflects the relative 3D positions of different humans. Skeletons of different individuals usually have similar lengths. We define the actual length of the human skeleton as $L$ and its length within the image as $l$. The pinhole camera model (Fusiello 2006), visualized in Fig. 6, gives us the relationship $L/l = D/f$, where $D$ is the depth and $f$ is the camera's focal length.

The dimension $l$ can be deduced from the human pose, with $L$ typically being a fixed value. Therefore, we propose that it is viable to utilize the pose to calculate the ratio $D/f$. Note that although the above reasoning uses $L$, neural networks do not require a predefined $L$. Instead, neural networks can learn skeleton lengths by using training datasets that include $D/f$ ratios and 2D human poses.

As detailed in Section 3.2, we adjust the raw image so that its longer edge measures 1,000 pixels. Denoting $S$ as the length of the raw image's long edge, and in accordance with the pinhole camera model, this resizing is equivalent to scaling the focal length $f$ to $1000f/S$.

For the training convenience, we introduce a constant coefficient $\alpha$ to normalize the measure within a [0,1] range. Finally, we compute the depth measure as:

$$\widetilde{D} = \frac{DS}{1000\alpha f} , \tag{3}$$

where $\alpha$ is set to 200 in practice. During training, we utilize the $L_1$ loss to supervise $\widetilde{D}$,

$$\mathcal{L}_{\mathrm{D}} = \|\widetilde{D} - \widetilde{D}^*\|_1 , \tag{4}$$

where $*$ represents the ground truth.

During the training phase, for datasets that include root-camera depth and focal length, we can directly compute the ground truth value for $\widetilde{D}$. However, in cases where these parameters are not provided, we assume the focal length, $f$, is 1500. Under this assumption, we use SMPLify (Pavlakos et al., 2019) or NeuralAnnot (Moon et al., 2022) to generate a pseudo root-camera depth, such that pseudo $\widetilde{D}$ can be obtained.

Assuming a constant focal length is sufficient for deriving a comprehensive depth. Because $f$ is the same for all humans within a single image, relative 3D body positions are consistently represented, irrespective of variations in $f$. During the inference stage, we can recover $D$ using $f$, $S$, and the predicted value of $\widetilde{D}$. If $f$ is not provided, we assume $f = 1500$.

### 3.5 Reconstructing Multiple Humans

In addition to depth estimation, our graph network produces, for each human, root-relative mesh vertex coordinates denoted as $M \in \mathbb{R}^{V\times 3}$ and 3D joint coordinates represented as $P^{3D} \in \mathbb{R}^{J\times 3}$. The coordinate matrix $M$ is essential for recovering the body mesh, and $P^{3D}$ plays a supporting role during training. Furthermore, we can derive the 3D joint coordinates directly from $M$ by multiplying the joint regression matrix $\mathcal{J}$. In the training phase, we use the $L_1$ loss function to supervise $M$, $P^{3D}$, and $\mathcal{J}M$,

$$\mathcal{L}_{\mathrm{M}} = ||M - M^{*}||_1 \ , \tag{5}$$

$$\mathcal{L}_{\mathrm{J}} = ||P^{3\mathrm{D}} - P^{3\mathrm{D}*}||_1 \ , \tag{6}$$

$$\mathcal{L}_{\mathrm{JM}} = ||\mathcal{J}M - P^{3\mathrm{D}*}||_1 \ , \tag{7}$$

where $*$ represents the ground truth. If the dataset lacks 3D labels for joints or mesh coordinates, we use SMPLify (Pavlakos et al., 2019) or NeuralAnnot (Moon et al., 2022) to generate pseudo 3D coordinates as a substitute for ground truth.

These losses, combined with the previously mentioned depth losses, are summed together to obtain the total loss,

$$\mathcal{L} = \mathcal{L}_{\mathrm{M}} + \lambda_1\mathcal{L}_{\mathrm{J}} + \lambda_2\mathcal{L}_{\mathrm{JM}} + \lambda_3\mathcal{L}_{\mathrm{D}} \ . \tag{8}$$

By optimizing these losses, we obtain accurate estimates of $M$ and $D$ for each human. Because we can directly obtain the root-joint 2D image coordinates $[X, Y]$ from the 2D pose input, the 3D coordinate for the root-joint $[x, y, D]$ can be computed using the camera intrinsic equation,

$$D[X, Y, 1]^T = I[x, y, D]^T \ , \tag{9}$$

where $I$ is the camera intrinsic matrix. In scenarios lacking explicit camera parameters, we construct the intrinsic matrix by assuming a focal length $f$ of 1,500 and designating the center point of the image as the principal point. Denoting the image size as $(W, H)$, Equation 9 becomes:

$$D\begin{bmatrix} X \\ Y \\ 1 \end{bmatrix} = \begin{bmatrix} f & 0 & \frac{W}{2} \\ 0 & f & \frac{H}{2} \\ 0 & 0 & 1 \end{bmatrix}\begin{bmatrix} x \\ y \\ D \end{bmatrix} \tag{10}$$

Finally, we can construct multiple humans by calculating the absolute body mesh coordinates for each individual:

$$M_{\mathrm{abs}} = M + [x, y, D] \ . \tag{11}$$

## 4 Experiments

We conducted a thorough evaluation of the capabilities of the MUG framework through a series of comprehensive experiments, designed to answer three pivotal questions:

- Can MUG accurately estimate 3D human depth relationships solely from 2D pose input?
- How does MUG's performance compare with existing multi-human 3D mesh reconstruction techniques across a variety of benchmark datasets?
- What specific roles do individual components of MUG, such as the inter-human edges and 2D pose inputs, play in its overall effectiveness?

To address these questions, we initially outline the evaluation metrics and datasets employed in our experiments in Section 4.1 and 4.2. This is followed by a detailed discussion of our implementation strategies in Section 4.3. Subsequently, we present comparative analyses with SOTA methods in Section 4.4 to respond to the first two questions. Then, to validate the individual contributions of MUG's components, we conduct an ablation study detailed in Section 4.5. Finally, Section 4.6 provides some qualitative examples.

### 4.1 Evaluation Metrics

- Percentage of Correct Depth Relations ($\mathrm{PCDR}^{0.2}$): This metric evaluates the accuracy of depth estimation by calculating the percentage of correctly predicted depth relationships between pairs of subjects within a threshold of 0.2.
- Mean Per Joint Position Error (MPJPE): This metric measures the average Euclidean distance between the predicted and ground-truth 3D joint locations after aligning the root joints.
- Procrustes-Aligned Mean Per Joint Position Error (PA-MPJPE): Similar to MPJPE, but the predicted and ground-truth 3D joint locations are first aligned using Procrustes analysis to remove differences due to rotation, translation, and scale.
- Mean Per Vertex Position Error (MPVPE): This metric computes the average Euclidean distance between the predicted and ground-truth 3D mesh vertex positions after aligning the root joints.
- 3D Percentage of Correct Keypoints (3DPCK): This metric calculates the percentage of predicted 3D keypoints that fall within a specified threshold distance from the ground-truth keypoints.

### 4.2 Datasets

- **Human3.6M** (Ionescu et al., 2013) is a large-scale single-human 3D pose dataset where the ground truth 3D poses were captured through motion capture in controlled indoor environments. Due to its high-quality 3D annotations, we use this dataset for both training and testing. However, as Human3.6M lacks body mesh ground truth, we adopt the pseudo-ground-truth provided by NeuralAnnot (Moon et al., 2022). We follow the evaluation protocol established by Kanazawa et al. (2018). We employ two metrics for evaluation: MPJPE and PA-MPJPE.
- **RH** (Sun et al., 2022) is an in-the-wild dataset designed for assessing depth perception, containing approximately 7,600 images featuring weakly annotated figures of more than 24,800 individuals. The dataset utilizes the Percentage of Correct Depth Relations ($PCDR^{0.2}$) with a 0.2m threshold as the standard for measuring depth reasoning accuracy.
- **COCO** (Lin et al., 2014) is a large-scale, in-the-wild, multi-human dataset, providing 2D pose ground truth. Although it lacks 3D pose ground truth, we incorporate COCO in our training to introduce greater diversity. We use pseudo human-mesh ground truth provided by NeuralAnnot (Moon et al., 2022).
- **MuCo-3DHP** is a multi-human 3D pose dataset proposed by Mehta et al. (2018). It is composed from the existing MPI-INF-3DHP 3D single-human pose estimation dataset (Mehta et al., 2017). We only use this dataset during the training phase. We adopt the pseudo-ground-truth provided by (Choi et al., 2020), where SMPLify (Pavlakos et al., 2019) is used to generate pseudo-ground-truth.
- **MuPoTS-3D** is an outdoor multi-human 3D pose dataset proposed by Mehta et al. (2018), where ground truth 3D pose is obtained by a multi-view, marker-less motion capture system. We use this dataset for evaluation purposes. We follow the standard evaluation method from (Jiang et al., 2020). The evaluation metric is the percentage of correct 3D keypoints (3DPCK).
- **Panoptic** (Joo et al., 2015) is a multi-human dataset, featuring video clips collected within a panoptic environment. It has ground-truth 3D pose annotations. We follow the evaluation procedure by Zanfir et al. (2018a). Haggling, Mafia, Ultimatum, and Pizza are four activities used to test. MPJPE is used for evaluation.
- **3DPW** (Von Marcard et al., 2018) is a multi-human, in-the-wild dataset that includes full-body mesh annotations. Following the same protocol as (Kocabas et al., 2020), we use the test set for evaluation. It is important to note that the training set of 3DPW is not utilized at any stage in our training process. MPJPE, PA-MPJPE, and MPVPE are used for evaluation.
- **Human-Art** (Ju et al., 2023) provides a collection of art images encompassing various types of drawings. For this dataset, we only present qualitative results based on its images.

### 4.3 Implementation Details

All the codes used in the experiments were implemented with PyTorch (Paszke et al., 2019). The GCN part uses the PyG (PyTorch Geometric) library (Fey & Lenssen, 2019). To optimize computational efficiency, we have implemented a coarse body mesh model characterized by a human mesh consisting of 431 vertices, denoted as $V = 431$. This coarse mesh was generated by applying a sampling algorithm of Ranjan et al. (2018) to sub-sample the original mesh (which has 6890 vertices, as per the SMPL human mesh topology) down to 431 vertices. Following Lin et al. (2021), we utilized trainable Multi-Layer Perceptrons (MLPs) to upsample the coarse mesh back to its original structure. The GCN and MLPs are trained end-to-end.

In our feature construction for most experiments, we utilized the COCO joint set as defined by Lin et al. (2014) to represent human keypoints. Following Choi et al. (2020), we added the root-joint and neck-joint, determined as the midpoints of the left and right hips, and the left and right shoulders, respectively. This resulted in $J = 19$ joints for the COCO joint set. However, for the experiments in Section 4.4.5, we employed the Human3.6M joint set, where $J = 17$. In the context of employing the COCO joint set, according to Equations 1 and 2, the feature dimension for each joint or mesh node is defined as:

$$d = 2 + 2 \times 19 + 2 + 3 \times 11 = 75. \tag{12}$$

With $K$ representing the total number of humans, the dimensionality for the joint features is expressed as $K \times 19 \times 75$, and for the mesh features, $K \times 431 \times 75$.

In training, we added synthetic errors to the 2D pose ground truth as the input, following (Moon et al., 2019b; Choi et al., 2020), which effectively improved the network robustness against inaccuracies in pose estimation. For the implementation specifics, we set 256 multi-human graphs as one batch. The training duration was set to 15 epochs, with an initial learning rate of 0.001. We employed a cosine learning rate scheduler and utilized the AdamW optimizer. Our training process involved joint training on Human3.6M, COCO, and MuCo-3DHP simultaneously. We set the coefficients in Equation 8 as follows: $\lambda_1 = 1e{-3}$, $\lambda_2 = 1e{-3}$, and $\lambda_3 = 1$.

During the inference stage, we used BigDetection (Cai et al., 2022) to locate human figures within the image. Subse-quently, ViTPose (Xu et al., 2022) was used for the estimation of 2D poses.

**Table 1** Performance comparison on the RH dataset. The metric is PCDR$^{0.2}$ (%). Values presented in bold indicate the best performance.

| Method | Baby | Kid | Teen | Adult | All |
|---|---|---|---|---|---|
| 3DMPPE (Moon et al., 2019a) | 39.33 | 51.42 | 60.91 | 57.95 | 57.47 |
| CRMH (Jiang et al., 2020) | 34.74 | 48.37 | 59.11 | 55.47 | 54.83 |
| SMAP (Zhen et al., 2020) | 31.58 | 40.29 | 47.35 | 41.65 | 41.55 |
| ROMP (Sun et al., 2021) | 30.08 | 48.41 | 51.12 | 55.34 | 54.81 |
| BEV (Sun et al., 2022) | **60.77** | 67.09 | 66.07 | **69.71** | 68.27 |
| MUG | 48.72 | **71.10** | **74.15** | 69.50 | **69.87** |

**Table 2** Performance comparison on the 3DPW dataset. The units of MPJPE, PA-MPJPE, and MPVPE are mm. "Multi" indicates multi-human mesh estimation approaches. [a] denotes the utilization of extra datasets; [b] indicates results obtained using our implementation to omit the 3DPW training set. Values presented in bold indicate the best performance.

| Type | Method | Multi | MPJPE | PA-MPJPE | MPVPE |
|---|---|---|---|---|---|
| Image-based | CRMH (Jiang et al., 2020) | ✓ | 105.3 | 62.3 | 122.2 |
| | BMP (Zhang et al., 2021) | ✓ | 104.1 | 63.8 | 119.3 |
| | ROMP (Sun et al., 2021) | ✓ | 91.3 | 54.9 | 108.3 |
| | ROMP[a] (Sun et al., 2021) | ✓ | 85.5 | **53.3** | 103.1 |
| | BEV[a,b] (Sun et al., 2022) | ✓ | 87.1 | 53.7 | 102.5 |
| Pose-based | Pose2Mesh (Choi et al., 2020) | – | 88.9 | 58.3 | 106.3 |
| | PQ-GCN (Wang et al., 2022) | – | 89.2 | 58.3 | 106.4 |
| | GTRS (Zheng et al., 2022a) | – | 88.5 | 58.9 | 106.2 |
| | GATOR (You et al., 2023) | – | 87.5 | 56.8 | 104.5 |
| | MUG | ✓ | **84.6** | 56.9 | **101.4** |

### 4.4 Comparison with the State of the Art

This section presents a comparative analysis between our methodology and various SOTA techniques in the field. Unless otherwise specified, as detailed in Section 4.3, our training procedure utilized three datasets, and for inference, we incorporated BigDetection and ViTPose for 2D pose estimation. Given the vast array of current human mesh reconstruction methods, our focus was narrowed to multi-human mesh approaches that operate using single monocular images. This focus led us to intentionally exclude methods that are dependent on video or multi-view inputs for their operation. Furthermore, for datasets such as Human3.6M and 3DPW, which are applicable to single-human benchmarks as well, we broadened our comparative scope to include pose-based, single-human approaches.

#### 4.4.1 Results on RH

Our study evaluated the monocular depth reasoning capability of MUG using the RH dataset. In Table 1, we present a comparison of MUG against other leading multi-human pose or mesh methods. It is important to note that all methods included in this comparison are image-based, due to the scarcity of pose-based methods proficient in monocular depth reasoning. The RH dataset was not included in the initial multi-dataset training. To align with benchmark evaluation criteria, particularly with BEV (Sun et al., 2022), we fine-tuned MUG specifically on the RH training set for optimal performance. We adopted the piece-wise depth layer loss proposed by BEV in fine-tuning. However, unlike BEV, MUG does not integrate age-related data. The results, as shown in the table, reveal that MUG outperformed BEV by 1.6 points and achieved greater gains over other methods. In the "Baby" category, the performance of MUG was notably weaker compared to BEV. This discrepancy can be attributed to two primary factors. First, BEV's integration of age-related information, which is absent in MUG, contributes to its enhanced performance in this category. Second, the unique skeletal proportions of babies present a specific challenge. These distinct anatomical characteristics make it inherently difficult for MUG, which relies on skeletal data, to accurately estimate the depth of babies. Conversely, MUG exhibited significant superiority in the "Teen" and "Kid" categories, surpassing BEV by more than 8 and 4 points, respectively, and achieved comparable results in the "Adult" category. This led to an overall superior performance for MUG, underscoring the effectiveness of our approach. Despite relying solely on 2D pose inputs, MUG effectively estimated relative 3D positions, surpassing the performance of existing image-based methodologies. Further insights and a detailed ablation analysis on the RH dataset will be presented in Section 4.5.

**Table 3** 3DPCK (%) comparison on MuPoTS-3D. The columns "All" and "Matched" represent the accuracy for all predictions and the predictions matched with annotations, respectively. Values presented in bold indicate the best performance.

| Method | All | Matched |
|---|---|---|
| SMPLify-X (Pavlakos et al., 2019) | 62.8 | 68.0 |
| HMR (Kanazawa et al., 2018) | 66.1 | 70.9 |
| CRMH (Jiang et al., 2020) | 69.1 | 72.2 |
| BMP (Zhang et al., 2021) | 73.8 | 75.3 |
| ROMP (Sun et al., 2021) | 69.9 | 74.6 |
| BEV (Sun et al., 2022) | 70.2 | 75.2 |
| MUG | **78.0** | **78.9** |

#### 4.4.2 Results on 3DPW

We evaluated the multi-human reconstruction performance on the 3DPW test set, with the results presented in Table 2. Following the protocol established by Kocabas et al. (2020), we deliberately excluded the 3DPW training set from our training process. Consistent with prior research (Kocabas et al., 2020; Sun et al., 2021), we utilized ground truth bounding boxes for accurate localization of human subjects, as opposed to employing BigDetection. The table presents our comparative analysis against existing multi-human reconstruction approaches. Notably, our method demonstrated performance comparable to recent methods such as ROMP and BEV. ROMP's assessment was conducted under two distinct settings. The basic setting, which aligns closely with our training data, utilized three datasets: Human3.6M COCO, and MPI-INF-3DHP (Mehta et al., 2017). ROMP's enhanced setting, however, incorporated a broader range of datasets, totaling 11. Despite this, our results indicate that MUG remains competitive, even against ROMP's enhanced setting. While our PA-MPJPE metric showed a disadvantage, we achieved better results in both MPJPE and MPVPE metrics. Additionally, we have reimplemented BEV, omitting the 3DPW training set while retaining the other datasets used in the original study. This adaptation of BEV has led to our method outperforming it by 2.5 mm in MPJPE and 0.9 mm in MPVPE. Subsequent comparisons of MUG with recent pose-based approaches revealed a substantial advancement. For instance, MUG outperformed GATOR by approximately 3 mm in both MPJPE and MPVPE. The results clearly demonstrate that our method is not only effective in estimating human depth but also excels in human mesh reconstruction.

#### 4.4.3 Results on MuPoTS-3D

Our method was compared with SOTA 3D multi-human mesh reconstruction techniques using the MuPoTS-3D dataset, as detailed in Table 3. Similar to our evaluations with 3DPW, this analysis also spanned across datasets, with MuPoTS-3D not being included in our training data. As shown in Table 3, MUG demonstrated a significant performance advantage over other SOTA methods. Specifically, when compared to the recent notable work BEV by Sun et al. (2022), our method surpassed it in both all and matched cases, as detailed in Table 3. These results affirm the strong generalization ability of MUG.

#### 4.4.4 Results on Panoptic

We examined the performance on the Panoptic dataset, a typical lab-collected dataset, where images were captured in a laboratory environment. The comparison results are presented in Table 4. Our method outperformed several well-known multi-human mesh approaches, including ROMP and BMP. In comparison to BEV, our method exhibited a slightly lower mean performance. However, it's noteworthy that we achieved leading results in specific categories, particularly "Haggling" and "Mafia".

To gain a deeper understanding of our approach's limitations, we conducted an in-depth analysis of failure cases. We provide an example in Fig. 7. A key issue identified is the camera angle in these indoor settings, often leading to only partial visibility of subjects. The pose network, in its current configuration, constrains keypoint predictions to fall within the image. This constraint can lead to significant errors, particularly in estimating the position of knees and ankles, which are often predicted inaccurately and far from their actual positions. Consequently, starting from these compromised pose predictions, our MUG model exhibited notable inaccuracies in reconstructing lower body parts. We recognize the potential for improving our model's performance on the Panoptic dataset if the pose network is capable of accurately predicting joints not visible within the image frame. Such an enhancement would significantly refine our approach. However, addressing this challenge lies outside the scope of this paper and is identified as a direction for future research. We anticipate that advancements in this area will greatly improve the efficacy of our model, particularly in handling the complexities of indoor imaging environments.

#### 4.4.5 Results on Human3.6M

Beyond our focus on multi-human benchmarks, we expanded our evaluation to include the performance of MUG on the Human3.6M dataset. Despite Human3.6M being a single-human dataset and our primary focus being on multi-human issues, the results from Human3.6M are instrumental in demonstrating MUG's proficiency in body mesh recovery. Our goal is to present a thorough and comprehensive overview of the human mesh estimation field to the reader. For this series of experiments, we followed the Pose2Mesh setting. Training was conducted using five subjects (S1, S5, S6, S7, and S8), while testing involved two different subjects (S9 and S11). During testing, the 2D pose information was provided by Pose2Mesh. The outcomes are presented in Table 5. When compared to earlier multi-human mesh methods, including BMP, our method achieved a 5.5% reduction in MPJPE. This improvement is attributed to the robust generalization capability of the 2D pose and the efficiency of our graph architecture, which effectively models intra-human relations. Moreover, in comparison to pose-based single-human methods such as Pose2Mesh and GATOR, our MUG method demonstrated superior performance in MPJPE, validating the advancement of our intra-human structure. The table also includes performance metrics for recent image-based single-human approaches, containing METRO and CLIFF. We acknowledge a notable performance gap between our method and SOTA for Human3.6M. This gap is understandable, considering that these single-human approaches are highly optimized for Human3.6M, often utilizing a broad spectrum of datasets for training, whereas our training was solely based on the Human3.6M dataset. Additionally, single-human approaches, which focus exclusively on individual subjects without addressing multi-human scenarios, tend to achieve better performance on single-human benchmarks due to their specialized nature. Similar gaps were observed even in SOTA multi-human methods like BEV. In this context, the performance of single-human approaches can serve as a benchmark for the potential upper limits of multi-human approaches.

**Table 4** Performance comparison on the Panoptic dataset. The evaluation metric is MPJPE. The unit is mm. Values presented in bold indicate the best performance.

| Method | Haggling | Mafia | Ultimatum | Pizza | Mean |
|---|---|---|---|---|---|
| MSC (Zanfir et al., 2018a) | 140.0 | 165.9 | 150.7 | 156.0 | 153.4 |
| MubyNet (Zanfir et al., 2018b) | 141.4 | 152.3 | 145.0 | 162.5 | 150.3 |
| CRMH (Jiang et al., 2020) | 129.6 | 133.5 | 153.0 | 156.7 | 143.2 |
| BMP (Zhang et al., 2021) | 120.4 | 132.7 | 140.9 | 147.5 | 135.4 |
| ROMP (Sun et al., 2021) | 111.8 | 129.0 | 148.5 | 149.1 | 134.6 |
| BEV (Sun et al., 2022) | **90.7** | **103.7** | 113.1 | 125.2 | **109.5** |
| MUG | 109.1 | 122.7 | **108.5** | **118.4** | 117.2 |

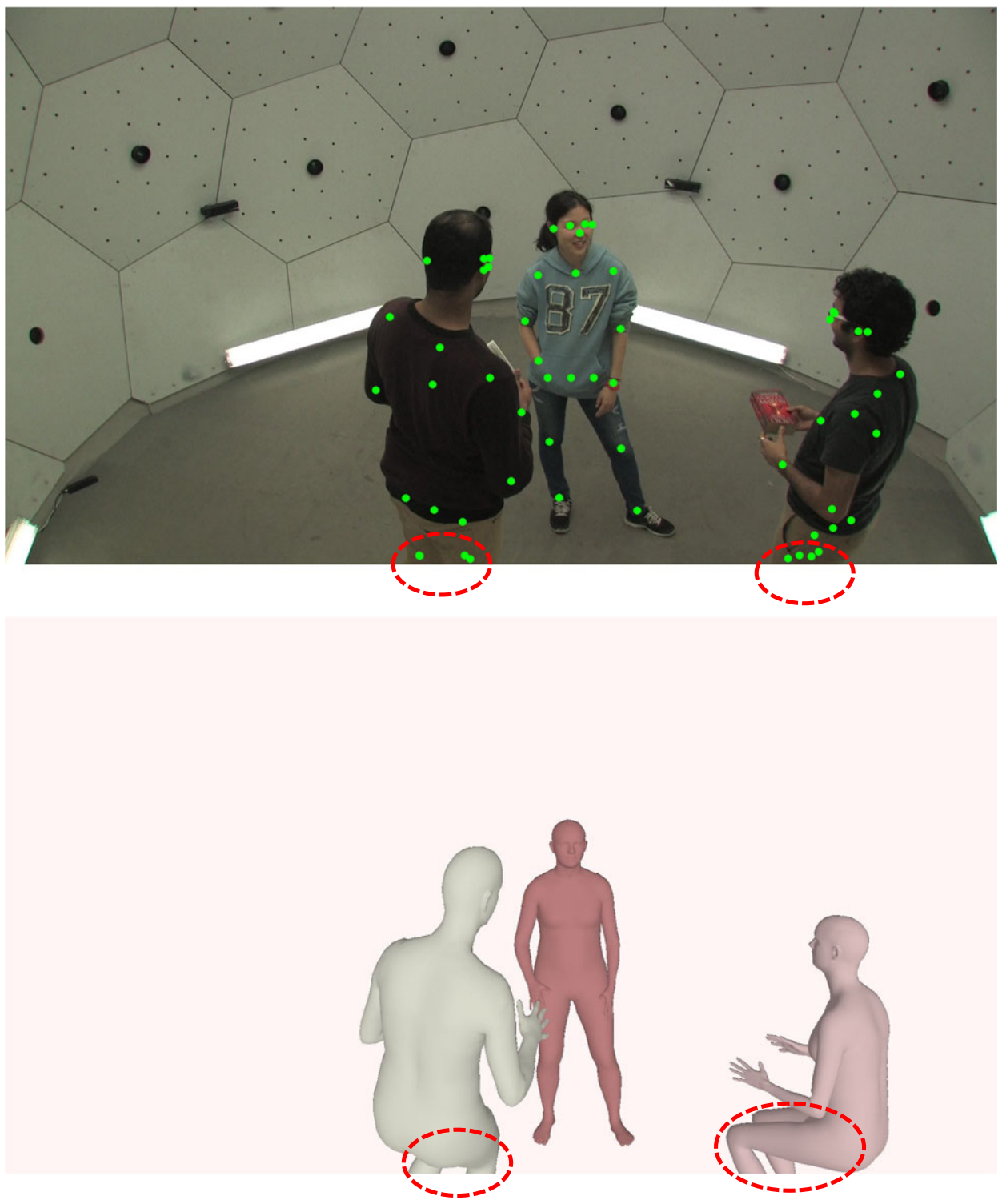

**Fig. 7** A failure case in the Panoptic dataset. In the upper image, the green points are the predicted joints from ViTPose. The lower image shows the human mesh reconstruction result obtained from MUG.

## 4.5 Ablation Analysis

This section presents the ablation studies conducted on MUG, with evaluations carried out using the 3DPW and RH test sets.

### 4.5.1 Ablation on the Network Structure

A key innovation in our study is the integration of a novel heterogeneous graph within the network architecture. This graph comprises two distinct types of nodes: mesh nodes and joint nodes. This configuration sets our method apart from previous methods like Pose2Mesh, which typically utilizes homogeneous graphs. To validate the effectiveness of this design, we conducted a series of experiments as detailed in Table 6. We explored three different structure variations: (a) using only mesh nodes and mesh branch, (b) implementing joint nodes and employing upsampling to convert these into mesh nodes, with the output being mesh nodes, and (c) ours. The results clearly indicate that ours significantly outperforms both (a) and (b), thereby affirming the advantages of a heterogeneous graph design that synergistically integrates both human skeleton and mesh graph structures throughout the network. Additionally, the results from Table 2 demonstrate that MUG outperforms Pose2Mesh

**Table 5** Performance comparison on the Human3.6M dataset. "Multi" indicates multi-human mesh estimation approaches. The units of MPJPE and PA-MPJPE are mm.

| Type | Method | Multi | MPJPE | PA-MPJPE |
|---|---|---|---|---|
| Image-based | GraphCMR (Kolotouros et al., 2019b) | – | 88.0 | 56.8 |
| | HybrIK (Li et al., 2021) | – | 54.4 | 34.5 |
| | METRO (Lin et al., 2021) | – | 54.0 | 36.7 |
| | CLIFF (Li et al., 2022) | – | 47.1 | 32.7 |
| | CRMH (Jiang et al., 2020) | ✓ | – | 52.7 |
| | BMP (Zhang et al., 2021) | ✓ | – | 51.3 |
| Pose-based | Pose2Mesh (Choi et al., 2020) | – | 64.9 | 46.3 |
| | PQ-GCN (Wang et al., 2022) | – | 64.6 | 47.9 |
| | GTRS (Zheng et al., 2022a) | – | 64.3 | 45.4 |
| | GATOR (You et al., 2023) | – | 64.0 | 44.7 |
| | MUG | ✓ | 59.3 | 47.2 |

**Table 6** Ablation for the graph construction on 3DPW.

| Graph Construction | MPVPE |
|---|---|
| Mesh Node → Mesh Node | 103.9 |
| Joint Node → Mesh Node | 104.2 |
| Joint&Mesh Node → Joint&Mesh Node | 101.4 |

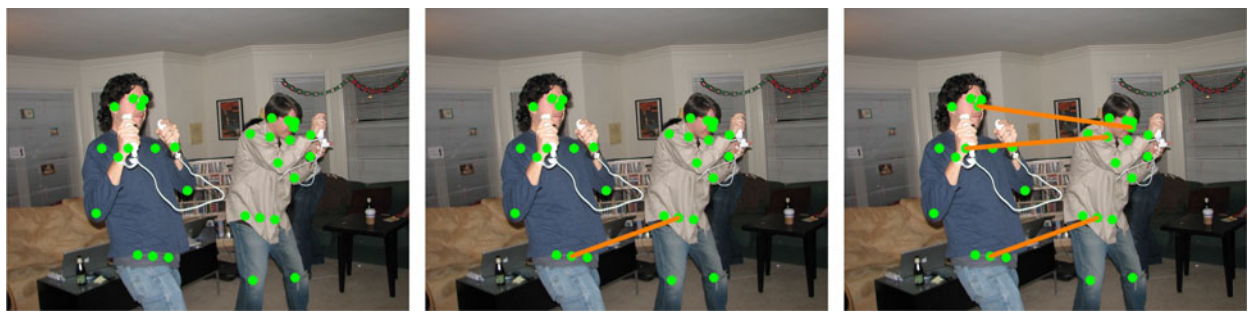

**Fig. 9** Illustration of inter-human edges, with the number of edges per human pair set at 0, 1, and 3, respectively. Green points represent body joints. Orange lines represent inter-human edges.

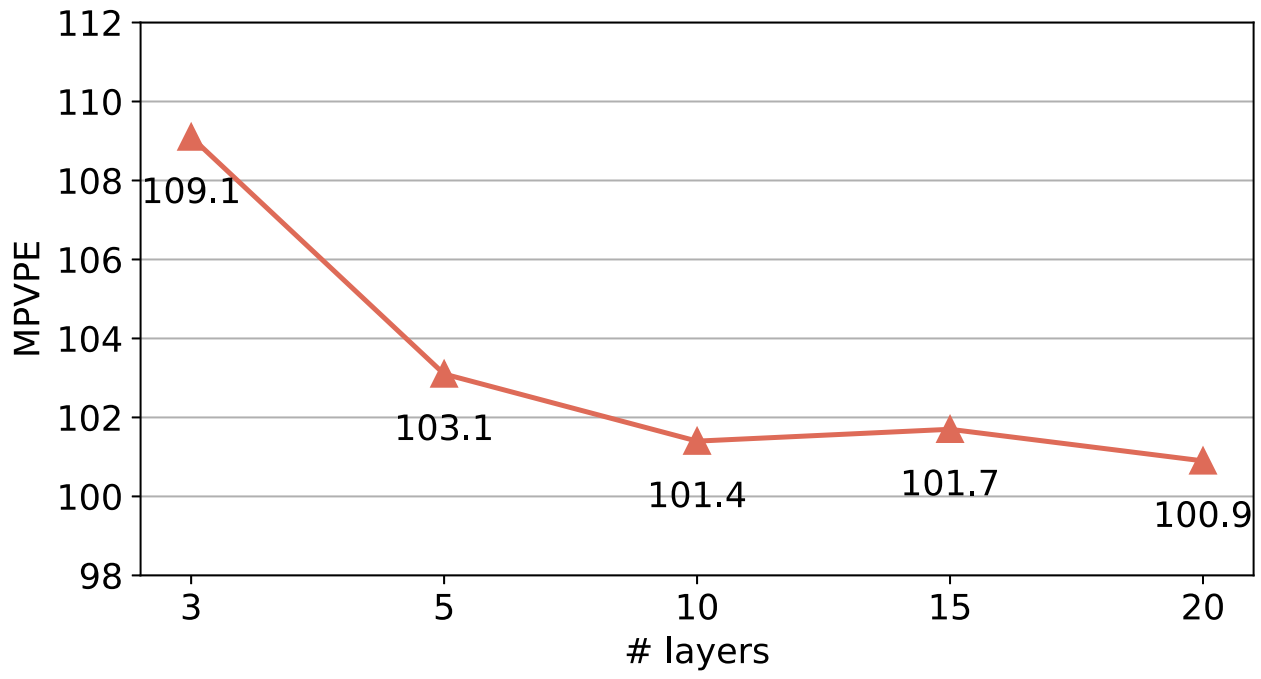

**Fig. 8** Ablation for the number of layers on 3DPW.

which adopts design (b), thereby supporting the validity of our conclusion.

Subsequently, we investigated the effect of the length of the graph network. For this purpose, we conducted experiments using varying network lengths of 3, 5, 10, 15, and 20 layers in both the pose and mesh branches. We regulated the number of layers by modifying the sequence of consecutive GCN residual blocks. Fig. 8 illustrates how the performance on the 3DPW dataset varies in relation to these different network lengths. Our findings suggest that a 10-layer network offers the most effective balance between performance and computational cost. Notably, extending the network beyond this length does not yield significant performance enhancements.

### 4.5.2 Ablation on Inter-human Edges

Our method is specifically designed for multi-human scenarios, with a key feature being our proposed graph structure, inter-human edges $\mathcal{E}^{\text{Inter}}$, which facilitate connections between different human subgraphs. We conducted an in-depth analysis to understand the impact of these inter-human edges on performance, focusing on human depth order and reconstruction results. We explored three different conditions: First, edges across different humans were omitted, rendering the graph as a collection of isolated subgraphs. Second, inter-human edges were established at the root-joints between different humans. Third, these inter-human connections were expanded to include the neck and nose joints. The number of inter-human edges per human pair is 0, 1, and 3 for the three conditions, respectively. Fig. 9 illustrates the three conditions. Table 7 presents findings on human depth ordering We trained the model on three datasets and then directly applied it to the RH dataset, along with fine-tuning the RH dataset. Our results indicate that connections between root joints yield the best outcomes. Notably, the absence of inter-human edges leads to a significant drop in performance, as it hinders the transmission of information necessary for analyzing relative 3D positions across different humans. Additionally, Table 8 reports the human reconstruction results on 3DPW. Here, it was observed that root joint connections do not significantly influence reconstruction out-comes. In fact, an increased number of connections may adversely affect performance.

**Table 7** Ablation for inter-human edges on RH. The metric is PCDR$^{0.2}$ (%).

| # edges per human pair | Fine-tuning | Baby | Kid | Teen | Adult | All |
|---|---|---|---|---|---|---|
| 0 | – | 28.79 | 47.16 | 55.57 | 54.73 | 54.80 |
| 1 | – | 31.37 | 50.29 | 59.94 | 57.11 | 57.37 |
| 3 | – | 29.53 | 48.33 | 57.57 | 56.70 | 56.56 |
| 0 | ✓ | 50.19 | 68.69 | 70.69 | 67.52 | 67.95 |
| 1 | ✓ | 48.72 | 71.10 | 74.15 | 69.50 | 69.87 |
| 3 | ✓ | 52.41 | 69.32 | 71.96 | 68.79 | 69.06 |

**Table 8** Ablation for inter-human edges on 3DPW.

| # edges per human pair | MPVPE |
|---|---|
| 0 | 101.4 |
| 1 | 101.6 |
| 3 | 102.4 |

**Table 9** Performance on the 3DPW dataset with different 2D pose inputs.

| 2D Pose Input | Jitter $\sigma$ (px) | MPVPE |
|---|---|---|
| HRNet | – | 103.9 |
| DarkPose+HRNet | – | 102.8 |
| ViTPose | 0 | 101.4 |
| ViTPose | 2 | 102.2 |
| ViTPose | 4 | 103.1 |
| ViTPose | 8 | 104.9 |
| ViTPose | 16 | 108.2 |
| Ground Truth | – | 69.4 |

### 4.5.3 Ablation on 2D Pose Input

We study how test-time 2D pose quality affects MUG. Although we inject synthetic pose errors during training (Moon et al., 2019b; Choi et al., 2020) to improve robustness, variation in the accuracy of input poses still impacts inference. On the 3DPW test set (MPVPE), we evaluate three estimators of increasing strength-HRNet (Sun et al., 2019), HRNet+DarkPose (Zhang et al., 2020), and ViTPose (Xu et al., 2022). As Table 9 shows, better 2D poses yield better meshes (HRNet → DarkPose+HRNet → ViTPose). Using ground-truth poses provides an upper bound, reducing MPVPE by 31.6% (from 101.4 mm to 69.4 mm), indicating substantial headroom as pose estimators improve.

We further assess robustness by adding i.i.d. Gaussian jitter to ViTPose keypoints and sweeping $\sigma_{\text{px}} \in \{2, 4, 8, 16\}$ (applied per joint to $(x, y)$; the third channel remains unchanged). Results in Table 9 show graceful degradation: MPVPE increases from 101.4 mm at $\sigma$=0 to 102.2/103.1/104.9/108.2 mm at $\sigma$=2/4/8/16 px, respectively. This confirms that MUG remains stable under small pose noise and degrades predictably under larger perturbations.

**Table 10** Ablation on overlap severity. The metrics for RH and 3DPW are PCDR$^{0.2}$ (%) and MPVPE, respectively.

| Overlap Level | 3DPW | RH |
|---|---|---|
| Low | 100.7 | 55.6 |
| Moderate | 102.0 | 53.5 |
| Heavy | 105.6 | 49.1 |

### 4.5.4 Ablation on Overlap Severity

We further evaluate robustness under different levels of inter-person overlap. Since dense occlusion annotations are unavailable on our evaluation sets, we use 2D overlap as a measurable proxy for overlap severity. For two subjects with 2D boxes $B_i$ and $B_j$, we define the overlap ratio as $o_{ij} = |B_i \cap B_j| / \min(|B_i|, |B_j|)$. Unless otherwise noted, the boxes are derived from the same person annotations used for evaluation, rather than from MUG predictions, so that the overlap bins reflect scene difficulty instead of model output quality. We divide the test samples into three bins: low overlap ($0 \le o_{ij} \le 0.1$), moderate overlap ($0.1 < o_{ij} \le 0.3$), and heavy overlap ($o_{ij} > 0.3$). On RH, each annotated pair is assigned its pairwise overlap and we report PCDR$^{0.2}$. On 3DPW, each subject is assigned the maximum overlap with any other subject in the same image, and we report MPVPE.

Table 10 suggests a consistent performance drop as overlap increases. On 3DPW, the MPVPE rises from 100.7 mm under low overlap to 102.0 mm under moderate overlap and 105.6 mm under heavy overlap. On RH, the PCDR$^{0.2}$ decreases from 55.6% to 53.5% and further to 49.1%. These results suggest that MUG is reasonably robust under low-to-moderate overlap, while heavy overlap makes both depth prediction and mesh reconstruction more difficult for a pose-only input.

**Table 11** Ablation for the feature construction on 3DPW and RH. The metrics for 3DPW and RH are MPVPE and $\mathrm{PCDR}^{0.2}$, respectively.

| $\overline{P}^{2D}_{ij}$ | $\widetilde{P}^{2D}_{ij}$ | $\widetilde{P}^{2D}_{i}$ | $(\mathcal{J}T)_j$ | 3DPW | RH |
|---|---|---|---|---|---|
| ✓ | | | | 107.1 | 34.8 |
| ✓ | ✓ | | | 105.0 | 52.8 |
| ✓ | ✓ | ✓ | | 102.9 | 55.2 |
| ✓ | ✓ | ✓ | ✓ | 101.4 | 54.8 |

**Table 13** Ablation on pseudo-ground-truth for selected training sets (Human3.6M and COCO). Evaluation conducted on the 3DPW dataset.

| Methods of Pseudo-Ground-Truth Mesh | MPVPE |
|---|---|
| SMPLify | 103.1 |
| NeuralAnnot | 101.4 |

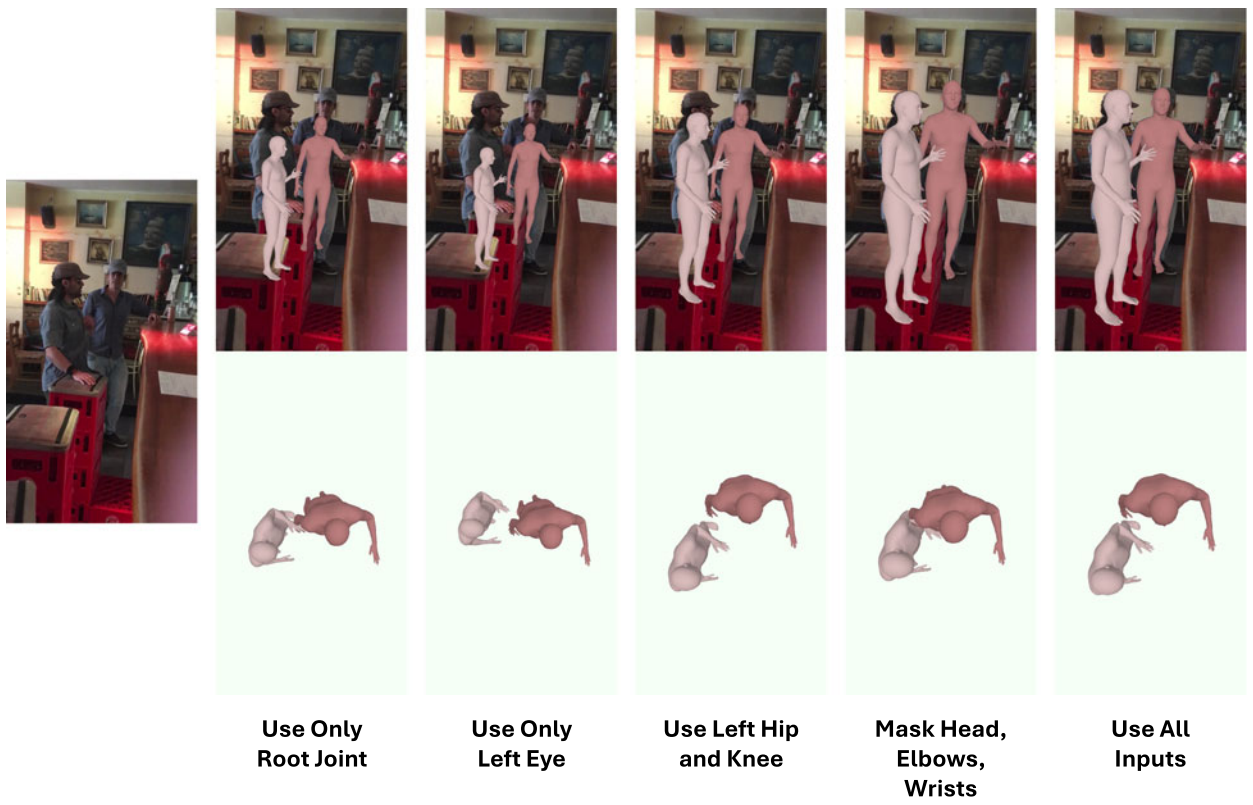

**Fig. 10** Impact of Specific Joints on Depth Estimation. The leftmost image is the input image. From left to right, the depth estimations of the human meshes are provided by the models listed in Table 12, starting from the row "Use Only Root Joint"to the final model.

### 4.5.5 Ablation on Node Feature Construction

As shown in Equations 1 and 2, The node features in our model are composed of four components: $\widetilde{P}^{2D}_{ij}$,$\widetilde{P}^{2D}_{i}$, $\overline{P}^{2D}_{ij}$, and $(\mathcal{J}T)_j$. Ablation experiments were conducted by directly applying the respective trained models on the 3DPW and RH datasets, illustrating the impact of each component. Table 11 reports the result. It is clear that each component contributes to the enhancement of the final result. Notably, the $\mathrm{PCDR}^{0.2}$ on the RH dataset, when utilizing only the normalized pose $\overline{P}^{2D}_{ij}$, registers at 34.8%. However, this performance markedly improves to 52.8% with the addition of the absolute pose $\widetilde{P}^{2D}_{ij}$, thereby emphasizing the vital importance of absolute pose in accurate 3D position inference.

### 4.5.6 Ablation on the Impact of Specific Joints on Depth Estimation

In Section 4.5.5, we demonstrate that absolute pose can aid in estimating 3D depth over normalized pose. In this section, we delve deeper into the impact of specific joints on depth estimation. We masked certain joints in the input. We trained our model on Human3.6, COCO, and MuCo-3DHP datasets, and evaluated it on the RH dataset.

Table 12 reports the quantitative results and masking strategies. When all joints are masked, the model is unable to estimate the relative depth of humans. Surprisingly, when only the root joint is used as input, the model can still estimate a reasonable relative depth. This suggests that the absolute position in the image already contains depth information. One explanation is that humans closer to the camera are more likely to appear lower in the image. The performance on the "Baby" and "Kid" categories is even better than with the full set of input joints. However, not all joints perform as well as the root joint. For example, as shown in Table 12, the performance using the left eye as input is poor. Using more joints as input, generally results in reasonable relative depth estimation.

However, using only the root joint is not sufficient to estimate the actual depth of humans. Figure 10 provides examples that combine the human mesh obtained from our final model with the estimated depth from each experiment. When using only the root joint, the relative depth order is correct. But the model cannot learn the actual size of the human body or place the human at the correct absolute distance, given the focal length. With more joints as input, the model can estimate the human size and position more accurately.

**Table 12** Impact of specific joints on relative depth estimation. The metric is $\mathrm{PCDR}^{0.2}$ (%). Performance metrics for different distance relations (equal, close, far) are provided.

| Masking Strategy | Baby | Kid | Teen | Adult | Equal | Close | Far | All |
|---|---|---|---|---|---|---|---|---|
| Mask All Joints | 43.55 | 31.62 | 23.87 | 28.53 | 100.0 | 0.0 | 0.0 | 28.91 |
| Use Only Root Joint | 39.12 | 59.76 | 56.84 | 56.09 | 15.43 | 77.6 | 71.89 | 55.88 |
| Use Only Left Eye | 43.37 | 38.32 | 37.71 | 33.7 | 62.35 | 25.25 | 23.67 | 34.81 |
| Use Left Hip and Knee | 33.03 | 50.38 | 61.21 | 56.48 | 18.24 | 82.42 | 59.36 | 56.58 |
| Mask Head, Elbows, Wrists | 29.89 | 47.7 | 54.11 | 55.55 | 16.48 | 82.59 | 55.09 | 55.32 |
| Use All Inputs | 31.37 | 50.29 | 59.94 | 57.11 | 18.30 | 85.27 | 56.31 | 57.37 |

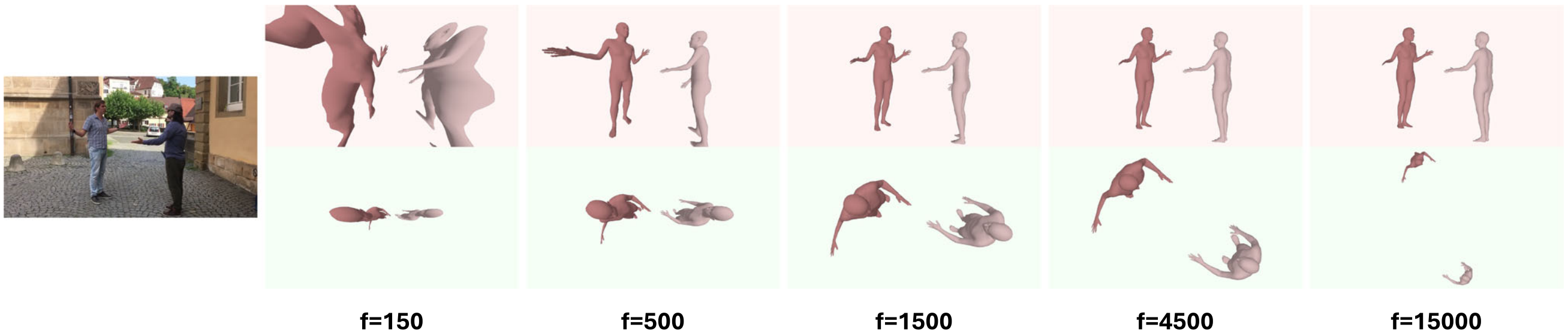

**Fig. 11** Qualitative comparison with different focal length selections. The front and top views at different focal lengths are shown.

**Table 14** Ablation for focal length on RH. The metric is PCDR$^{0.2}$ (%). Performance metrics for different distance relations (equal, close, far) are provided.

| Focal Length | Baby | Kid | Teen | Adult | Equal | Close | Far | All |
|---|---|---|---|---|---|---|---|---|
| 150 | 40.60 | 54.04 | 49.37 | 54.34 | 81.23 | 55.06 | 21.93 | 54.31 |
| 500 | 32.85 | 53.51 | 61.03 | 58.81 | 42.93 | 78.55 | 45.95 | 59.24 |
| 1500 | 31.37 | 50.29 | 59.94 | 57.11 | 18.30 | 85.27 | 56.31 | 57.37 |
| 4500 | 30.08 | 47.52 | 57.93 | 55.68 | 5.98 | 87.03 | 59.97 | 55.37 |
| 15000 | 28.79 | 47.34 | 57.57 | 55.16 | 1.52 | 87.83 | 61.36 | 54.75 |

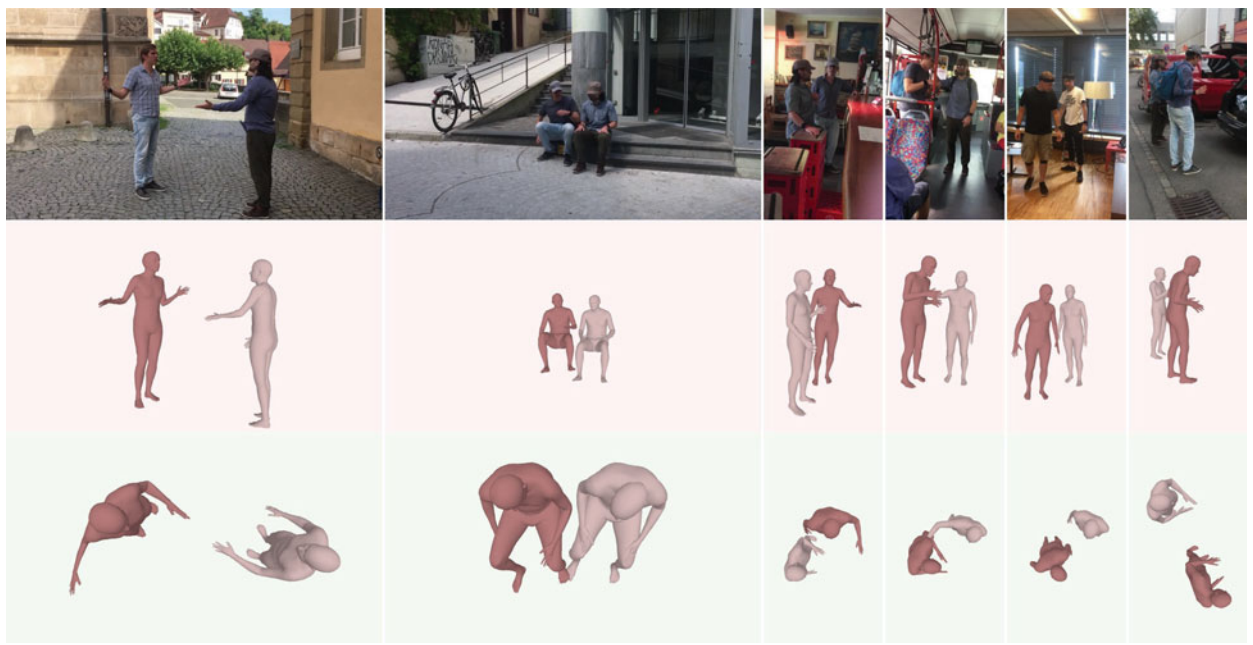

**Fig. 12** Qualitative results on 3DPW. The front and top views are shown.

#### 4.5.7 Ablation on Pseudo-ground-truth

We use pseudo-ground-truth human body meshes in training. Specifically, the pseudo-ground-truth for MuCo-3DHP is generated by SMPLify (Pavlakos et al., 2019). Meanwhile, for Human3.6M and COCO, we use the pseudo-ground-truth provided by NeuralAnnot (Moon et al., 2022), as NeuralAnnot does not provide pseudo-ground-truth for MuCo-3DHP. Table 13 shows that using NeuralAnnot for Human3.6M and COCO in training results in a 1.7 mm MPVPE improvement. This is because NeuralAnnot provides more accurate human mesh coordinates compared to SMPLify.

#### 4.5.8 Ablation on Focal Length Selection

As mentioned in Section 3.4 and 3.5, during inference, in scenarios lacking explicit camera parameters, we assume $f = 1500$. Since $f$ is the same for all humans within a single image, relative 3D body positions are consistently represented regardless of $f$. However, according to equation 3, the actual distance $D$ is directly proportional to $f$. Therefore, a smaller $f$ implies a shorter root-camera distance, which in turn reduces the relative depth across humans. Here, we analyze the effect of $f$ on human reconstruction output in inference.

We provide qualitative results in Fig. 11. As shown, a small $f$ can cause distortion in humans from the front view, and the relative depth appears reduced. Conversely, a larger $f$ results in greater distances.

Quantitative results are presented in Table 14 for the RH dataset, which sets the threshold for equal depth at 0.2 meters. A smaller $f$ tends to result in more predictions being classified as equal depth. In contrast, a larger $f$ leads to predictions being classified as either close or far. $f = 500$ provides the best quantitative performance, with more accurate equal depth predictions compared to $f = 1500$.

$f = 1500$ represents a trade-off between qualitative human reconstruction results and quantitative performance.

### 4.6 Qualitative Results

Figures 12 and 13 present qualitative results of MUG on 3DPW, RH, Panoptic, and MuPoTS-3D. The impressive qualitative results demonstrate MUG's robust capability to handle images from diverse sources, irrespective of background and appearance variations, thereby underscoring its strong generalization ability.

Furthermore, we have applied MUG to photos of artworks from the Human-Art dataset. Paintings frequently present unique challenges to computer vision algorithms, primarily due to their nature as artistic abstractions rather than direct representations of reality, as is the case with pho-tographs. Unlike photographs that capture the real world with precision, paintings often embody the artist's subjective interpretation and stylistic expression, which can vary significantly from the literal appearance of their subjects. Fig. 14 shows that SOTA like BEV often struggle with artistic images due to the domain adaptation challenge between their training data and such novel scenarios. In contrast, MUG maintains its effectiveness thanks to robust 2D pose network. Given accurate pose estimates, MUG consistently generates high-quality human meshes across diverse image domains.

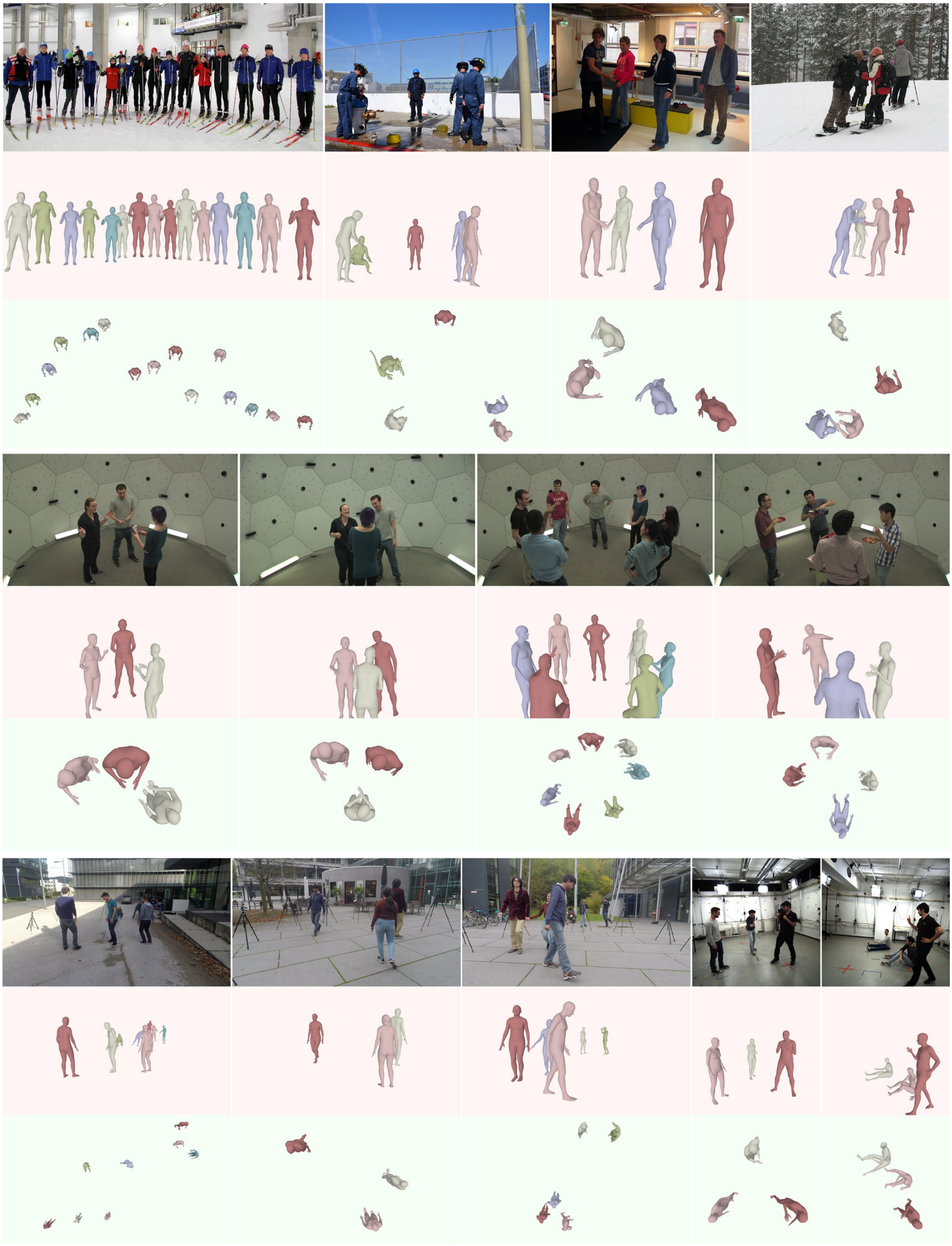

**Fig. 13** Qualitative results on RH, Panoptic, and MuPoTS-3D. Three rows of original images are sourced from RH, Panoptic, and MuPoTS-3D, respectively. The front and top views are shown.

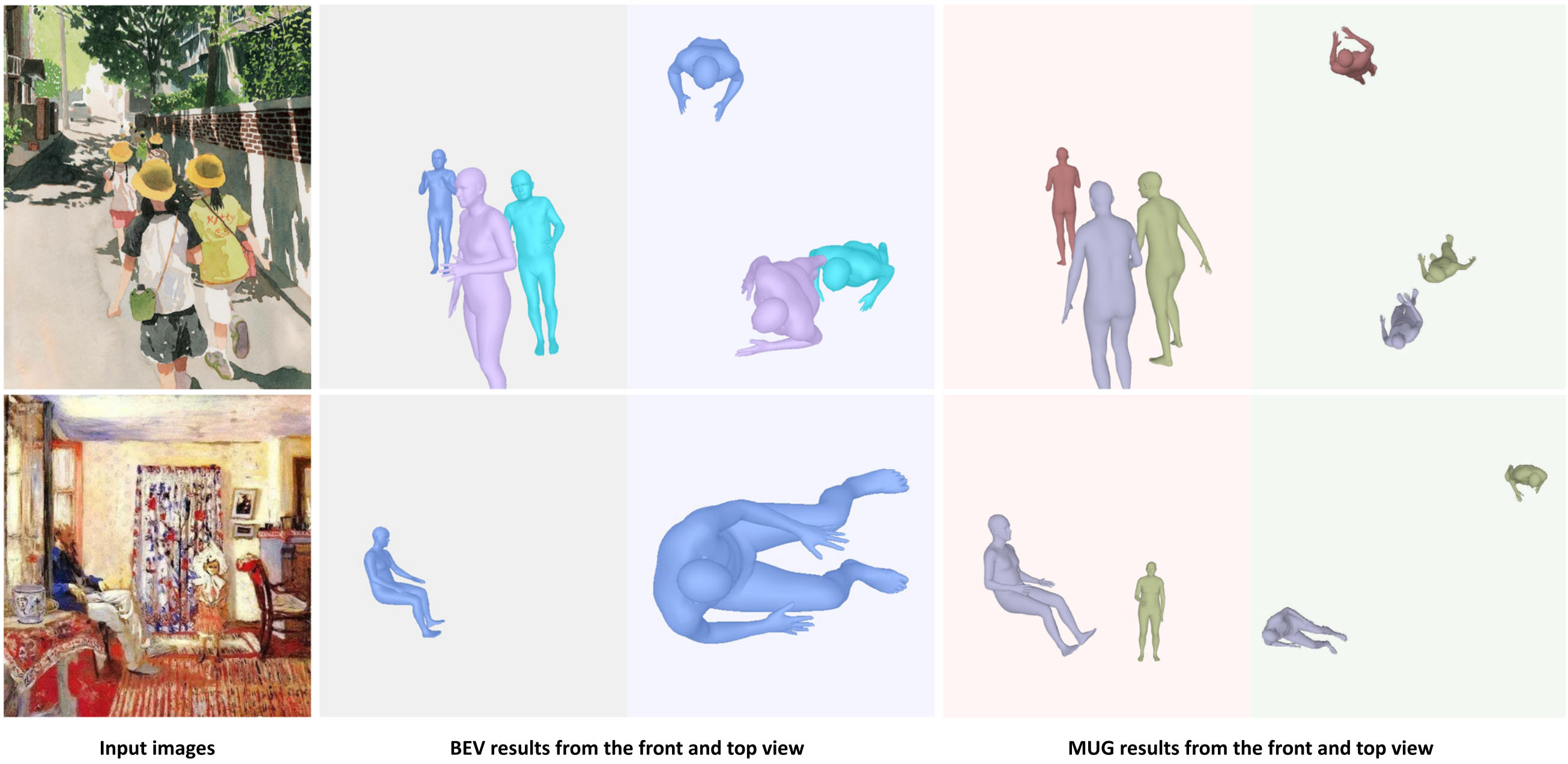

**Fig. 14** Qualitative comparison to the SOTA method BEV on Human-Art. First row: *Untitled*, watercolor, 2021, by Seong Ryul. Courtesy of HK ARTION. Second row: *The Painter Ker-Xavier Roussel and His Daughter*, oil on cardboard, 1903, by Edouard Vuillard. Courtesy of The Buffalo AKG Art Museum, Buffalo, New York.

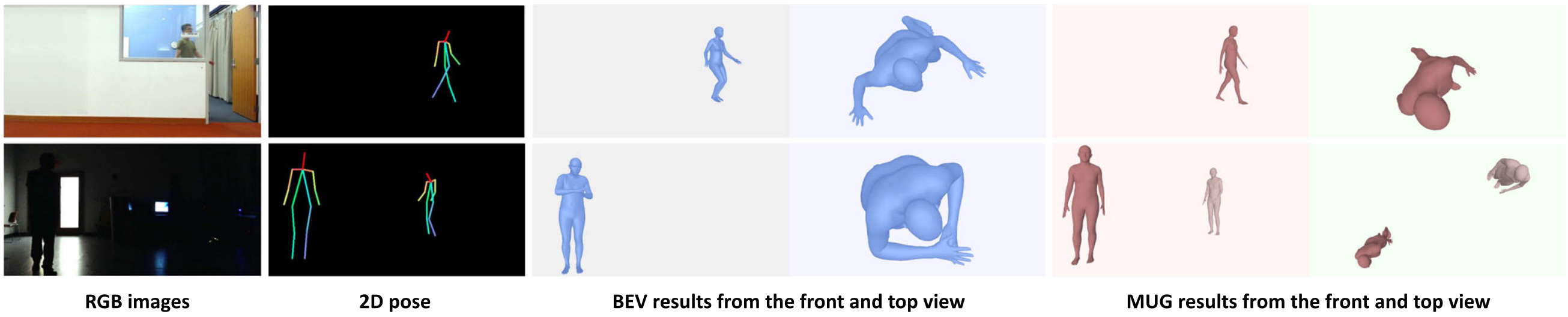

**Fig. 15** MUG effectively addresses through-wall and poor lighting scenarios by utilizing a 2D human pose network with radio signal inputs (Zhao et al., 2018). RGB images and estimated 2D pose images are sourced from Zhao et al. (2018). The 2D pose coordinate values were extracted from the 2D pose images using OpenCV and used as the input for MUG. For comparison, results from BEV, which uses RGB images as input, are also included.

Finally, we show how MUG leverages 2D pose estimates from radio signals to reconstruct human meshes in scenarios where traditional image-based methods struggle. Building upon the radio signal-based pose detection network proposed by Zhao et al. (2018), our method effectively handles challenging cases such as through-wall captures and low-illumination conditions. As illustrated in Fig. 15, image-based methods fail in these scenarios due to their lack of exposure to such conditions during training, producing incorrect lower body predictions under wall occlusion and failing to detect humans in poor lighting. In contrast, MUG accurately reconstructs human meshes using the keypoints estimated from radio signals. In the future, MUG could also collaborate with some LiDAR-based human pose networks, such as Zheng et al. (2022b), to address scenarios that are challenging for image-based approaches.

## 5 Conclusions and Future Work

We explored multi-human mesh reconstruction using 2D pose as the input. To achieve this, we designed a multi-layer heterogeneous graph that represents both intra-human and inter-human relations. Furthermore, we introduced a dual-branch graph neural network that employs this heterogeneous graph to effectively learn human relationships and accurately reconstruct 3D mesh coordinates. Our study shows that a pose-based pipeline not only effectively addresses the challenges in multi-person scenarios but also currently outperforms image-based methods across several datasets, showcasing robust generalization capabilities. We hope our findings encourage further research in this area, for example by incorporating image features into the graph representation.

Moreover, integrating 3D mesh models into the field of bodily expressed emotion understanding presents a promising direction for future work. Such integration could provide nuanced insights into the relationship between human physicality and emotional expression, paving the way for advancements in affective computing and related disciplines.

**Supplementary Information** The online version contains supplementary material available at https://doi.org/10.1007/s11263-026-02942-9.

**Acknowledgements** The authors appreciate the support and encouragement from Yelin Kim and Adam Fineberg. They also thank the anonymous reviewers and associate editor for their valuable feedback and constructive comments, which have significantly improved the quality of this manuscript.

**Author Contributions** Conceptualization: Chenyan Wu; Methodology: Chenyan Wu; Formal analysis and investigation: Chenyan Wu; Writing - original draft preparation: Chenyan Wu; Writing - review and editing: all authors; Funding acquisition: James Z. Wang; Resources: James Z. Wang; Supervision: James Z. Wang.

**Funding** This research was supported by generous gifts from the Amazon Research Awards program and by the National Science Foundation (NSF) under Grant No. CNS-2234195. This work used cluster computers at the Pittsburgh Supercomputer Center and the National Center for Supercomputing Applications through an allocation from the Advanced Cyberinfrastructure Coordination Ecosystem: Services & Support (ACCESS) program, which is supported by NSF Grants Nos. 2138259, 2138286, 2138307, 2137603, and 2138296. This work also used the Extreme Science and Engineering Discovery Environment, which was supported by NSF Grant No. 1548562.

**Data Availability** The datasets generated and/or analyzed during the current study are available in the Human3.6M repository, http://vision.imar.ro/human3.6m/description.php, the RH repository, https://github.com/Arthur151/Relative_Human, the COCO dataset, https://cocodataset.org/#home, the MuCo-3DHP repository, https://vcai.mpi-inf.mpg.de/projects/SingleShotMultiPerson/, the MuPoTS-3D repository, https://vcai.mpi-inf.mpg.de/projects/SingleShotMultiPerson/, the Panoptic repository, http://domedb.perception.cs.cmu.edu/, the 3DPW repository, https://virtualhumans.mpi-inf.mpg.de/3DPW/, and the Human-Art repository, https://idea-research.github.io/HumanArt/.

## Declarations

**Competing interests** The authors declare that they have no known competing financial interests or personal relationships that could have appeared to influence the work reported in this manuscript. This statement is made in the interest of full disclosure and to uphold the standards of ethical scholarly practice. Our research has been conducted with the highest level of integrity, and the results presented are a true and unbiased representation of our findings.

## References

Bogo, F., Kanazawa, A., Lassner, C., Gehler, P., Romero, J., & Black, M. J. (2016). Keep it SMPL: Automatic estimation of 3D human pose and shape from a single image. *Proceedings of the European Conference on Computer Vision* (pp. 561–578). Springer.

Cai, L., Zhang, Z., Zhu, Y., Zhang,L., Li,M., & Xue,X. (2022) BigDetection: A large-scale benchmark for improved object detector pre-training. arXiv preprint arXiv:2203.13249

Cao, Z., Simon, T., Wei, S. E., & Sheikh, Y. (2017). Realtime multi-person 2D pose estimation using part affinity fields. *Proceedings of the IEEE Conference on Computer Vision and Pattern Recognition* (pp. 7291–7299)

Choi, H., Moon, G., & Lee, K. M. (2020). Pose2Mesh: Graph convolutional network for 3D human pose and mesh recovery from a 2D human pose. *Proceedings of the European Conference on Computer Vision* (pp. 769–787). Springer.

Choi, H., Moon, G., Park, J., & Lee, K. M. (2022). Learning to estimate robust 3D human mesh from in-the-wild crowded scenes. *Proceedings of the IEEE/CVF Conference on Computer Vision and Pattern Recognition* (pp. 1475–1484)

Fey, M., & Lenssen, J. E. (2019). Fast graph representation learning with PyTorch Geometric. ICLR Workshop on Representation Learning on Graphs and Manifolds

Fusiello A (2006) Elements of geometric computer vision. https://homepages.inf.ed.ac.uk/rbf/CVonline/LOCAL_COPIES/FUSIELLO4/tutorial.html

Gong, K., Zhang, J., & Feng, J. (2021). PoseAug: A differentiable pose augmentation framework for 3D human pose estimation. *Proceedings of the IEEE/CVF Conference on Computer Vision and Pattern Recognition* (pp. 8575–8584)

Hamilton, W. L., Ying, R., & Leskovec, J. (2017). Inductive representation learning on large graphs arXiv preprint. arXiv:1706.02216

He, K., Gkioxari, G., Dollár, P., & Girshick, R. (2017). Mask R-CNN. In: Proceedings of the IEEE International Conference on Computer Vision, (pp. 2961–2969)

Ionescu, C., Papava, D., Olaru, V., & Sminchisescu, C. (2013). Human3.6M: Large scale datasets and predictive methods for 3D human sensing in natural environments. *IEEE Transactions on Pattern Analysis and Machine Intelligence,36*(7), 1325–1339.

Jiang, W., Kolotouros, N., Pavlakos, G., Zhou, X., & Daniilidis, K. (2020). Coherent reconstruction of multiple humans from a single image. *Proceedings of the IEEE/CVF Conference on Computer Vision and Pattern Recognition* (pp. 5579–5588)

Joo, H., Liu, H., Tan, L., Gui, L., Nabbe, B., Matthews, I., Kanade, T., Nobuhara, S., & Sheikh, Y. (2015). Panoptic studio: A massively multiview system for social motion capture. *Proceedings of the IEEE International Conference on Computer Vision* (pp. 3334–3342)

Ju, X., Zeng, A., Wang, J., Xu, Q., & Zhang, L. (2023). Human-art: A versatile human-centric dataset bridging natural and artificial scenes. *Proceedings of the IEEE/CVF Conference on Computer Vision and Pattern Recognition* (pp. 618–629)

Kanazawa, A., Black, M. J., Jacobs, D. W., & Malik, J. (2018). End-to-end recovery of human shape and pose. *Proceedings of the IEEE Conference on Computer Vision and Pattern Recognition* (pp. 7122–7131)

Ke, S. R., Thuc, H. L. U., Lee, Y. J., Hwang, J. N., Yoo, J. H., & Choi, K. H. (2013). A review on video-based human activity recognition. *Computers, 2*(2), 88–131.

Kipf, T. N., & Welling, M. (2016). Semi-supervised classification with graph convolutional networks arXiv preprint arXiv:1609.02907.

Kocabas, M., Athanasiou, N., & Black, M. J. (2020). Vibe: Video inference for human body pose and shape estimation. *Proceedings of the IEEE/CVF Conference on Computer Vision and Pattern Recognition* (pp. 5253–5263)

Kolotouros, N., Pavlakos, G., Black, M.J., & Daniilidis, K. (2019a). Learning to reconstruct 3D human pose and shape via model-fitting in the loop. In: Proceedings of the IEEE/CVF International Conference on Computer Vision, (pp. 2252–2261)

Kolotouros, N., Pavlakos, G., & Daniilidis, K. (2019b). Convolutional mesh regression for single-image human shape reconstruction. In: Proceedings of the IEEE/CVF Conference on Computer Vision and Pattern Recognition, (pp. 4501–4510)

Li, G., Xiong, C., Thabet, A.,& Ghanem, B. (2020). DeeperGCN All you need to train deeper GCNs. arXiv preprint arXiv:2006.07739

Li, J., Xu, C., Chen, Z., Bian, S., Yang, L., & Lu, C. (2021). Hybrik: A hybrid analytical-neural inverse kinematics solution for 3D human pose and shape estimation. *Proceedings of the IEEE/CVF International Conference on Computer Vision* (pp. 3383–3393)

Li, M., Chen, S., Chen, X., Zhang, Y.,Wang, Y.,& Tian, Q. (2019). Actional-structural graph convolutional networks for skeleton-based action recognition. In: Proceedings of the IEEE/CVF Conference on Computer Vision and Pattern Recognition, (pp. 3595–3603)

Li, Z., Liu, J., Zhang, Z., Xu, S.,& Yan, Y. (2022). Cliff: Carrying location information in full frames into human pose and shape estimation. In: Proceedings of the European Conference on Computer Vision, Springer, (pp. 590–606)

Lin, K., Wang, L., & Liu, Z. (2021). End-to-end human pose and mesh reconstruction with transformers. *Proceedings of the IEEE/CVF Conference on Computer Vision and Pattern Recognition* (pp. 1954–1963)

Lin, T. Y., Maire, M., Belongie, S., Hays, J., Perona, P., Ramanan, D., Dollár, P., & Zitnick, C. L. (2014). Microsoft COCO: Common objects in context. *Proceedings of the European Conference on Computer Vision* (pp. 740–755). Springer.

Loper, M., Mahmood, N., Romero, J., Pons-Moll, G., & Black, M. J. (2015). SMPL: A skinned multi-person linear model. *ACM Transactions on Graphics, 34*(6), 1–16.

Luo, Y., Ye, J., Adams, R. B., Li, J., Newman, M. G., & Wang, J. Z. (2020). ARBEE: Towards automated recognition of bodily expression of emotion in the wild. *International Journal of Computer Vision, 128*(1), 1–25. https://doi.org/10.1007/s11263-019-01215-y

von Marcard, T., Henschel, R., Black, M.J., Rosenhahn, B.,& Pons-Moll, G. (2018). Recovering accurate 3D human pose in the wild using imus and a moving camera. In: Proceedings of the European Conference on Computer Vision, (pp. 601–617)

Martinez, J., Hossain, R., Romero, J.,& Little, J.J. (2017). A simple yet effective baseline for 3D human pose estimation. In: Proceedings of the IEEE International Conference on Computer Vision, (pp. 2640–2649)

Mehta, D., Rhodin, H., Casas, D., Fua, P., Sotnychenko, O., Xu, W., & Theobalt, C. (2017). Monocular 3D human pose estimation in the wild using improved CNN supervision. *Proceedings of the International Conference on 3D Vision (3DV)* (pp. 506–516). IEEE.

Mehta, D., Sotnychenko, O., Mueller, F., Xu, W., Sridhar, S., Pons-Moll, G., & Theobalt, C. (2018). Single-shot multi-person 3D pose estimation from monocular RGB. *Proceedings of the International Conference on 3D Vision (3DV)* (pp. 120–130). IEEE.

Moon, G., & Lee, K. M. (2020). I2L-MeshNet: Image-to-lixel prediction network for accurate 3D human pose and mesh estimation from a single RGB image arXiv preprint. arXiv:2008.03713

Moon, G., Chang, J.Y.,& Lee, K.M. (2019a). Camera distance-aware top-down approach for 3D multi-person pose estimation from a single RGB image. In: Proceedings of the IEEE/CVF International Conference on Computer Vision, (pp. 10133–10142)

Moon, G., Chang, J.Y.,& Lee, K.M.(2019b). Posefix: Model-agnostic general human pose refinement network. In: Proceedings of the IEEE/CVF Conference on Computer Vision and Pattern Recognition, (pp. 7773–7781)

Moon, G., Choi, H., & Lee, K. M. (2022). NeuralAnnot Neural annotator for 3D human mesh training sets. *Proceedings of the IEEE/CVF Conference on Computer Vision and Pattern Recognition Workshop* (pp. 2299–2307)

Paszke, A., Gross, S., Massa, F., and others (2019). PyTorch: An imperative style, high-performance deep learning library. In: Advances in Neural Information Processing Systems 32. Curran Associates, Inc., (pp. 8024–8035), https://pytorch.org

Pavlakos, G., Zhu, L., Zhou, X., & Daniilidis, K. (2018). Learning to estimate 3D human pose and shape from a single color image. *Proceedings of the IEEE Conference on Computer Vision and Pattern Recognition* (pp. 459–468)

Pavlakos, G., Choutas, V., Ghorbani, N., Bolkart, T., Osman, A., Tzionas, D., & Black, M. J. (2019). Expressive body capture: 3D hands, face, and body from a single image. *Proceedings of the IEEE/CVF Conference on Computer Vision and Pattern Recognition* (pp. 10975–10985)

Ranjan, A., Bolkart, T., Sanyal, S., & Black, M. J. (2018). Generating 3D faces using convolutional mesh autoencoders. *Proceedings of the European Conference on Computer Vision* (pp. 704–720)

Ren, S., He, K., Girshick, R., & Sun, J. (2015). Faster R-CNN: Towards real-time object detection with region proposal networks. *Advances in Neural Information Processing Systems, 28*, 91–99.

Schlichtkrull, M., Kipf, T. N., Bloem, P., Van Den, Berg, R., Titov, I., & Welling, M. (2018). Modeling relational data with graph convolutional networks. *Proceedings of the European Semantic Web Conference* (pp. 593–607). Springer.

Sigal, L., Balan, A. O., & Black, M. J. (2010). Humaneva: Synchronized video and motion capture dataset and baseline algorithm for evaluation of articulated human motion. *International Journal of Computer Vision, 87*(1–2), 4–27.

Sun, K., Xiao, B., Liu, D., & Wang, J. (2019). Deep high-resolution representation learning for human pose estimation. *Proceedings of the IEEE/CVF Conference on Computer Vision and Pattern Recognition* (pp. 5693–5703)

Sun, Y., Bao, Q., Liu, W., Fu, Y., Black, M. J., & Mei, T. (2021). Monocular, one-stage, regression of multiple 3D people. *Proceedings of the IEEE/CVF International Conference on Computer Vision* (pp. 11179–11188)

Sun, Y., Liu, W., Bao, Q., Fu, Y., Mei, T., & Black, M. J. (2022). Putting people in their place: Monocular regression of 3D people in depth. *Proceedings of the IEEE/CVF Conference on Computer Vision and Pattern Recognition* (pp. 13243–13252)

Veličković, P., Cucurull, G., Casanova, A., Romero, A., Lio, P., & Bengio, Y. (2017). Graph attention networks arXiv:1710.10903.

Wandt, B., & Rosenhahn, B. (2019). RepNet: Weakly supervised training of an adversarial reprojection network for 3D human pose estimation. *Proceedings of the IEEE/CVF Conference on Computer Vision and Pattern Recognition* (pp. 7782–7791)

Wang, C., Li, J., Liu, W., Qian, C., & Lu, C. (2020). HMOR: Hierarchical multi-person ordinal relations for monocular multi-person 3D pose estimation. In: Proceedings of the European Conference on Computer Vision, Springer, (pp. 242–259)

Wang, J. Z., Zhao, S., Wu, C., Adams, R. B. J., Newman, M. G., Shafir, T., & Tsachor, R. (2023). Unlocking the emotional world of visual media: An overview of the science, research, and impact of understanding emotion. *Proceedings of the IEEE, 111*, 1–51. https://doi.org/10.1109/JPROC.2023.3273517

Wang, L., Chen, Y., Guo, Z., Qian, K.,Lin, M.,Li, H.,& Ren, J.S. (2019a). Generalizing monocular 3D human pose estimation in the wild. In: Proceedings of the IEEE/CVF International Conference on Computer Vision Workshops

Wang, L., Liu, X., Ma, X., Wu, J., Cheng, J., & Zhou, M. (2022). A progressive quadric graph convolutional network for 3D human mesh recovery. *IEEE Transactions on Circuits and Systems for Video Technology, 33*(1), 104–117.

Wang, X., Ji, H., Shi, C., Wang, B.,Ye, Y.,Cui, P.,& Yu, P.S. (2019b). Heterogeneous graph attention network. In: The World Wide Web Conference, (pp. 2022–2032)

Wu, C., Davaasuren, D., Shafir, T., Tsachor, R.,& Wang, J.Z. (2023). Bodily expressed emotion understanding through integrating laban movement analysis. *Patterns, 4*(10), 100816:1–14.

Xu, Y., Zhang, J., Zhang, Q., & Tao, D. (2022). ViTpose: Simple vision transformer baselines for human pose estimation. *Advances in Neural Information Processing Systems,35*, 38571–38584.

Yang, Y., Qiu, J., Song, M., Tao, D., & Wang, X. (2020). Distilling knowledge from graph convolutional networks. *Proceedings of the IEEE/CVF Conference on Computer Vision and Pattern Recognition* (pp. 7074–7083)

You, Y., Liu, H., Li, X., Li, W.,Wang, T.,& Ding, R. (2023). GATOR: Graph-aware transformer with motion-disentangled regression for human mesh recovery from a 2D pose. In: Proceedings of the IEEE International Conference on Acoustics, Speech and Signal Processing (ICASSP), IEEE, (pp. 1–5)

Zanfir, A., Marinoiu, E.,& Sminchisescu, C. (2018a). Monocular 3D pose and shape estimation of multiple people in natural scenes-the importance of multiple scene constraints. In: Proceedings of the IEEE Conference on Computer Vision and Pattern Recognition, (pp. 2148–2157)

Zanfir, A., Marinoiu, E., Zanfir, M., Popa, A. I., & Sminchisescu, C. (2018). Deep network for the integrated 3D sensing of multiple people in natural images. *Advances in Neural Information Processing Systems, 31*, 8410–8419.

Zhang, F., Zhu, X., Dai, H., Ye, M.,& Zhu, C. (2020). Distribution-aware coordinate representation for human pose estimation. In: Proceedings of the IEEE/CVF International Conference on Computer Vision, (pp. 7093–7102)

Zhang, J., Yu, D., Liew, J. H., Nie, X., & Feng, J. (2021). Body meshes as points arXiv preprint arXiv:2105.02467.

Zhao, M., Li, T., Abu Alsheikh, M., Tian, Y., Zhao, H.,Torralba, A.,& Katabi, D.(2018). Through-wall human pose estimation using radio signals. In: Proceedings of the IEEE/CVF Conference on Computer Vision and Pattern Recognition, (pp. 7356–7365)

Zhen, J., Fang, Q., Sun, J., Liu, W., Jiang, W., Bao, H., & Zhou, X. (2020). SMAP: Single-shot multi-person absolute 3D pose estimation. *Proceedings of the European Conference on Computer Vision* (pp. 550–566). Springer.

Zheng, C., Mendieta, M., Wang, P., Lu, A.,& Chen, C. (2022a). A lightweight graph transformer network for human mesh reconstruction from 2D human pose. In: Proceedings of the 30th ACM International Conference on Multimedia, (pp. 5496–5507)

Zheng, J., Shi, X., Gorban, A., Mao, J.,Song, Y.,Qi, C.R.,Liu, T.,Chari, V.,Cornman, A.,Zhou, Y., and others. (2022b) Multi-modal 3D human pose estimation with 2D weak supervision in autonomous driving. In: Proceedings of the IEEE/CVF Conference on Computer Vision and Pattern Recognition, (pp. 4478–4487)

Zhou, X., Huang, Q., Sun, X., Xue, X., & Wei, Y. (2017). Towards 3D human pose estimation in the wild: A weakly-supervised approach. *Proceedings of the IEEE International Conference on Computer Vision* (pp. 398–407)